\documentclass{article}

\usepackage{arxiv}

\usepackage{amsmath, amsfonts, amssymb, amsthm, amsbsy, amscd, bm, bbm,mathrsfs}
\usepackage[utf8]{inputenc} 
\usepackage[T1]{fontenc}    
\usepackage{hyperref}       
\usepackage{url}            
\usepackage{booktabs}       
\usepackage{amsfonts}       
\usepackage{nicefrac}       
\usepackage{microtype}      
\usepackage{graphicx}
\usepackage{subfig}
\usepackage{doi}
\usepackage[numbers]{natbib}
\usepackage[ruled,vlined]{algorithm2e}
\SetKw{kwInput}{Input:}
\SetKw{kwOutput}{Output:}
\SetKw{KwBreak}{break}

\title{MoR: Mixture of Representations for Mixed-Precision Training}

\author{
    Bor-Yiing Su\\
    Nvidia\\
    \texttt{boryiings@nvidia.com} \\
    \And
    Peter Dykas\\
    Nvidia\\
    \texttt{wdykas@nvidia.com} \\
    \And
    Mike Chrzanowski\\
    Nvidia\\
    \texttt{mchrzanowski@nvidia.com} \\
    \And
    Jatin Chhugani \thanks{Work done while at Nvidia.}\\
    Meta\\
    \texttt{jatinch@meta.com} \\
}

\date{}

\renewcommand{\headeright}{}

\begin{document}
\maketitle

\begin{abstract}

Mixed-precision training is a crucial technique for scaling deep learning models, but successful
mixed-precision training requires identifying and applying the right combination of training methods.
This paper presents our preliminary study on Mixture-of-Representations (MoR), a novel, per-tensor
and sub-tensor level quantization framework that dynamically analyzes a tensor’s numerical properties
to select between a variety of different representations. Based on the framework, we have proposed
and experimented concrete algorithms that choose dynamically between FP8 and BF16 representations for both
per-tensor and sub-tensor level granularities. Our universal approach is designed to preserve model
quality across various quantization partition strategies and datasets. Our initial findings show that
this approach can achieve state-of-the-art results with 98.38\% of tensors quantized to the FP8 format.
This work highlights the potential of dynamic, property-aware quantization while preserving model
quality. We believe this approach can generally improve the robustness of low precision training, as demonstrated by achieving FP8 accuracies that are on par with existing approaches without the need for fine-grain partitioning, or can be used in combination with other training methods to improve the leverage of even lower precision number formats such as NVFP4.

\end{abstract}


\section{Introduction}\label{sec:intro}

The scaling law \cite{scaling_law} suggests that model performance increases with the number of parameters and
the volume of training data, fueling a race to train ever-larger foundation models to achieve state-of-the-art
performance on industry benchmarks. For instance, the Llama 3.1 405B model was trained on 24,576 H100 GPUs for
months \cite{llama3}, the Megatron-Turing NLG 530B model on 4,480 GPUs for months \cite{Megatron530B}, the GPT-4
model was estimated to have been trained on over 20,000 GPUs for three months \cite{openai2023gpt4}, and the 
DeepSeek-V3 671B model is trained with 2.788M GPU hours \cite{DeepSeekV3}.
Consequently, training large foundation models is time-consuming, power-hungry, and expensive. This has led to
significant investment in optimizing the training workflow to accelerate performance and reduce resource consumption.
Among the most effective optimizations is the use of low-precision numerical representations.

Using low-precision representations offers numerous benefits: (a) From a compute perspective, modern hardware
offers significant speedups for lower-precision numerical operations. For example, an NVIDIA H100 GPU delivers
2x the FLOPS for FP8 GEMM operations compared to BF16 \cite{H100_arch}, while the newer GB200 and GU300 GPUs
provide 4x and 6x FLOPS for FP4 GEMM, respectively \cite{NVIDIA_GB200_NVL72, NVIDIA_GB300_Ultra}. (b) From a bandwidth
perspective, loading a vector of FP8 values requires only half the memory bandwidth of loading a BF16 vector of the
same length. Therefore, low-precision representations directly reduce the latency of both computation and memory
access, significantly accelerating training. (c) Moreover, applying low-precision formats during training enables
their direct use for inference, creating consistency between the two phases and obviating the need for
post-training quantization (PTQ) or quantization-aware training (QAT).

The formal specification of the FP8 format can be found in \cite{fp8_paulius}. 
It defines two FP8 formats: E4M3 and E5M2. The E4M3 format has 4 bits for exponent, and 3 bits for mantissa;
while the E5M2 format has 5 bits for exponent, and 2 bits for mantissa. 
But FP8 training is not only about the FP8 formats. In order to increase the representation accuracy,
improve the training stability, and also retain the model quality, many prior arts define
comprehensive recipes for FP8 training. \cite{delayed_scaling} and \cite{fp8_paulius} suggested current and delayed
per-tensor scaling. Essentially, defining a scaling factor for the full tensor to shift the numerical range
of the tensor to the representable range by the format type. Delayed scaling computes this scaling factor
based on historical absolute maximum (amax) values of the tensor; while current scaling computes this
scaling factor based on the amax of the current tensor. \cite{block_scaling} suggested partitioning
the tensor into blocks, and compute one scaling factor per block. Llama3 \cite{llama3} uses per-channel scaling that
computes one scaling factor per row or column based on the dot product direction. DeepSeekV3 \cite{DeepSeekV3} 
combines both sub-channel scaling and per-block scaling into the recipe, and proposes using
$1 \times 128$ sub-channel scaling for the activation tensor, and $128 \times 128$ per-block scaling for the 
weight tensor to achieve a balance between fine grained control and computational overhead. The micro-scaling format \cite{ocp_microscaling, microscaling, asit_mxfp8} suggested using a fine-grained $1 \times 32$ 
sub-channel block to compute the scaling factor. The NVFP4 format \cite{nvfp4} suggested using 
a fine-grained $1 \times 16$ sub-channel block to compute the scaling factor.

Another common strategy is to selectively apply FP8 quantization.  
\cite{sun2019hybrid, fp8_paulius} used E4M3 format in the forward pass, and uses E5M2 format in the backward pass.
\cite{coat} used per-group scaling for the optimizer states, and per-tensor scaling for the activation
tensors. \cite{fp4alltheway} suggested using fp4 format for the majority of the training, but
fall back to higher precision (FP32/BF16) at the late stage of training. \cite{nvidia_nvfp42025} applied nvfp4 for most of the layers, but left the first two and last eight layers in BF16, it also studied how reducing the number of BF16 layers affect model quality.

All of these prior arts focus on selecting the right methods and determining their appropriate application point (location and time) to create a low-precision training recipe that preserves model quality.
What we are trying to accomplish, however, is to start gaining insights into why a specific recipe can achieve model quality on par with the baseline. Therefore, we developed the MoR framework, which attempts to capture invariance through tensor analysis. In this paper, we also showed that relative error can serve as a reasonable invariance for FP8 training. While works like DeepSeekV3 \cite{DeepSeekV3} and MXFP8 \cite{asit_mxfp8} achieve good results because their fine-grained partition ensures that the quantization error is small for all tensors, we illustrated that using relative error as an invariance allows us to obtain on-par model quality with much coarser-grained partitions.
The relative error invariance we use is conservative and serves as a sufficient, but not a necessary, condition. We plan to continue working in this direction to explore invariance metrics that are more efficient and can be used for more aggressive quantization formats (such as NVFP4).

Our primary contributions are as follows:
\begin{enumerate}
  \item We propose a new strategy for computing the scaling factor that ensures consistency across the entire
  tensor, regardless of the chosen scaling granularity.
  \item We introduce a novel dynamic quantization framework that makes real-time decisions based on numerical properties at runtime, 
  allowing the model to adapt precision throughout the training process.
  \item Using this framework, we develop and evaluate tensor-level and sub-tensor-level \textbf{Mixture-of-Representations (MoR)} 
  recipes that dynamically select FP8 and BF16 types, achieving substantial efficiency without compromising quality.
\end{enumerate}

\section{Group Amax Mantissa Scaling}\label{sec:amax_scaling}

A major challenge for low-precision quantization is that it requires mapping a tensor's wide dynamic range 
to the limited representable range of the target format, such as FP8. 
For example, the E4M3 FP8 format can only represent positive values between $2^{-9}$ and 448; and the 
E5M2 FP8 format can only represent positive values between $2^{-16}$ and 57,344.
Quantizing values outside this range leads to information loss through either 
\textbf{saturation} (clipping large values) or \textbf{underflow} (flushing small values to zero). 
An effective strategy is to compute a scaling factor that
shifts the tensor's values into the representable range, 
minimizing these quantization errors and preserving numerical fidelity.

The prevailing hypothesis in quantization is that large-magnitude 
values are more critical to preserve than small ones. Consequently, 
the standard scaling strategy, used in all prior arts \cite{fp8_paulius, delayed_scaling, coat, block_scaling, llama3, DeepSeekV3, microscaling}, 
is to compute a scaling factor that maps the absolute maximum value (amax) 
of a tensor to the maximum representable value of the quantization format. 
This prioritizes avoiding saturation at the cost of flushing smaller values to zero.

Beyond the strategy, the numerical format of the scaling factor itself 
introduces a critical trade-off. Using a high-precision format like FP32 preserves the 
amax value the best but adds significant metadata overhead. 
To reduce this cost, prior work has explored lower-precision formats. 
The micro-scaling format \cite{microscaling} uses E8M0, which offers a 
wide dynamic range for the scaling factor, while NVFP4 \cite{nvfp4} suggests E4M3, 
which represents the amax with higher accuracy but has a more limited range. 
These approaches present a fundamental conflict: one must choose 
between the wider \textbf{range} of E8M0 and the higher \textbf{precision} of E4M3,
or falling back to FP32 with higher memory cost.

To resolve this trade-off, we propose the \textbf{Group Amax Mantissa (GAM) scaling algorithm}. 
The goal of GAM is to combine the respective strengths of prior approaches: 
to design a scaling factor representation that achieves the wide dynamic range 
of an E8M0-like exponent while preserving the tensor's maximum value with the 
high precision characteristic of an FP32 mantissa.

The core idea of GAM, detailed in Algorithm~\ref{algo:gam_scaling}, is to decouple 
the mantissa and exponent of the scaling factors. We partition a tensor $\mathbf{X}$ 
into blocks $b$, which are organized into groups $g$. For each group, 
we find the group-level amax ($g_{amax}$) and compute an ideal FP32 scaling factor $s_g$. 
We then store only the mantissa ($m_g$) of this group-level scaling factor. 
Subsequently, for each block $b$ within that group, we compute its local amax ($b_{amax}$) 
and its corresponding ideal FP32 scaling factor $s_b$. From $s_b$, we extract only the exponent. 
A crucial rounding step adjusts this block exponent downward if the block's 
mantissa ($m_b$) is smaller than the group's shared mantissa ($m_g$). 
This is to prevent saturation from happening. 
To apply scaling, the per-block FP32 scaling factor is 
reconstructed on-the-fly by combining the shared 23-bit 
group mantissa $m_g$ with the stored 8-bit per-block exponent.

While the group concept is generic, there is a trade-off between overhead and quantization error.
Smaller groups will have more overhead but with smaller quantization error, while larger groups
will have less overhead but with larger quantization error.
In our experiments, we use a single group for the entire tensor,
which is sufficient for the FP8 format as shown in our experiments in Section \ref{sec:exp_tensor}. This
configuration offers three primary benefits:
\begin{enumerate}
    \item \textbf{Negligible Overhead:} The storage cost is minimal. Each block requires an 8-bit exponent,
    and the entire tensor only needs one additional 23-bit mantissa.
    \item \textbf{Maximum Precision:} The absolute maximum value of the entire tensor is used to derive
    the mantissa, preserving it with full FP32 precision.
    \item \textbf{Consistent Mantissa Operations:} During scaling or de-scaling steps, the mantissa component of the reconstructed
    scaling factor is identical for all values. This makes the mantissa operations orthogonal to
    the block sizes. On the other hand, the exponent value of the scaling factor still depends
    on the block size and is subject to change based on the block amax value.
\end{enumerate}

\begin{algorithm}[t]
\KwIn{A tensor $\mathbf{X}$; a partition $G = \{g_i\}$ over the elements of $\mathbf{X}$; for each group $g \in G$, a partition $B_g = \{b_j\}$ of $g$ into blocks, quantization type amax $q_{amax}$}
\KwOut{The union of all group mantissas $M$ and all block scale factors in E8M0 $S$}

\ForEach{group $g$ in $G$}{
    $g_{amax} \leftarrow max(abs(g))$\;
    $s_g \leftarrow q_{amax} / g_{amax}$\;
    $m_g \leftarrow mantissa(s_g)$\;
    
    \ForEach{block $b$ in $B_g$}{
        $b_{amax} \leftarrow max(abs(b))$\;
        $s_b \leftarrow q_{amax} / b_{amax}$\;
        $m_b \leftarrow mantissa(s_b)$\;
        \tcc{Preventing saturation from happening by rounding down the exponent when $m_g > m_b$}
        \If{$m_g <= m_b$}{
            $s_b \leftarrow exponent(s_b)$\;
        }
        \Else{
            $s_b \leftarrow exponent(s_b) - 1$\;
        }
    }
}
\KwRet{$M \leftarrow \bigcup_{g \in G} \{m_g\}$, $S \leftarrow \bigcup_{b \in B} \{s_b\}$}
\caption{Group Amax Mantissa Scaling}
\label{algo:gam_scaling}
\end{algorithm}

\section{The MoR Framework}\label{sec:mor_framework}

The Mixture-of-Representations (MoR) framework, outlined in Algorithm~\ref{algo:mor_framework}, provides a
systematic approach for dynamically selecting quantization strategies. Given an input tensor, the framework
first partitions it into a set of blocks $B$ according to a chosen quantization granularity. For per-tensor
quantization, $B$ contains a single block (the entire tensor); for 2D block quantization, $B$ is the set
of all 2D blocks; for per-channel quantization, $B$ consists of the corresponding
rows or columns based on the dot product dimension; for sub-channel quantization, $B$ is the union of all sub-channel 
sub-rows or sub-columns.

The framework operates by iterating through an ordered list of target quantization types, ranked from \textbf{most aggressive(e.g. E4M3) to least aggressive (e.g. BF16)}.
For each block, the goal is to successfully apply the most preferred low-precision format. A set
of predefined metrics determines whether a given quantization type is acceptable for the block. If the
acceptance criteria for the current type are met, the block is quantized using that type, and the process
concludes for that block. If the criteria are not met, the framework falls back to the next type in the
list. In the [E4M3, E5M2, BF16] example, if the metric for E4M3 returns false, the framework then evaluates E5M2. If E5M2 also
fails, it falls back to BF16, effectively leaving the block in its original precision. When using the GAM
scaling algorithm from Section~\ref{sec:amax_scaling}, the additional metadata $A$ 
in the algorithm refers to the group mantissa $M$, which can be used in the metrics evaluations.
The block quantization metadata $D$ refers to the E8M0 block scaling factor $S$,
which is one of the output values when quantizing the block.

This approach is similar to the one presented in the FGMP paper \cite{fgmp}, which uses per-channel
quantization and independently quantizes each row or column into FP4 or FP8 based on metrics such as
quantization error, estimated matrix multiplication error, or block sensitivity.

Building on our MoR framework, we developed and evaluated two distinct mixed-precision recipes:
tensor-level MoR and sub-tensor-level MoR, which are detailed in the following sections.

\begin{algorithm}[t]
\KwIn{A tensor $\mathbf{X}$; a partition $B = \{b_i\}$ of $\mathbf{X}$ into blocks; an ordered list of tensor representation types $T_1$ to $T_k$; an ordered list of criteria metrics $M_1$ to $M_{k - 1}$; additional metadata $A$ required to perform quantization }
\KwOut{The quantized tensor $\mathbf{X_Q}$ and all block quantization metadata $D$}

\ForEach{block $b$ in $B$}{
    \For{$i \leftarrow 1$ \KwTo $k$}{
        \If{$i = k$}{
            $b_Q, d_b \leftarrow$ Quantize block $b$ to type $T_k$\;
            \KwBreak\;
        }
        \If{$M_i(B, b, A)$}{
            $b_Q, d_b \leftarrow$ Quantize block $b$ to type $T_i$\;
            \KwBreak\;
        }
    }
}

\KwRet{$\mathbf{X_Q} \leftarrow \bigcup_{b \in B} \{b_Q\}$, $D \leftarrow \bigcup_{b \in B} \{d_b\}$}
\caption{MoR Framework}
\label{algo:mor_framework}
\end{algorithm}

\subsection{MoR at Tensor Granularity}\label{sec:tensor}

\begin{figure}[htbp]
    \centering
    \includegraphics[width=\textwidth]{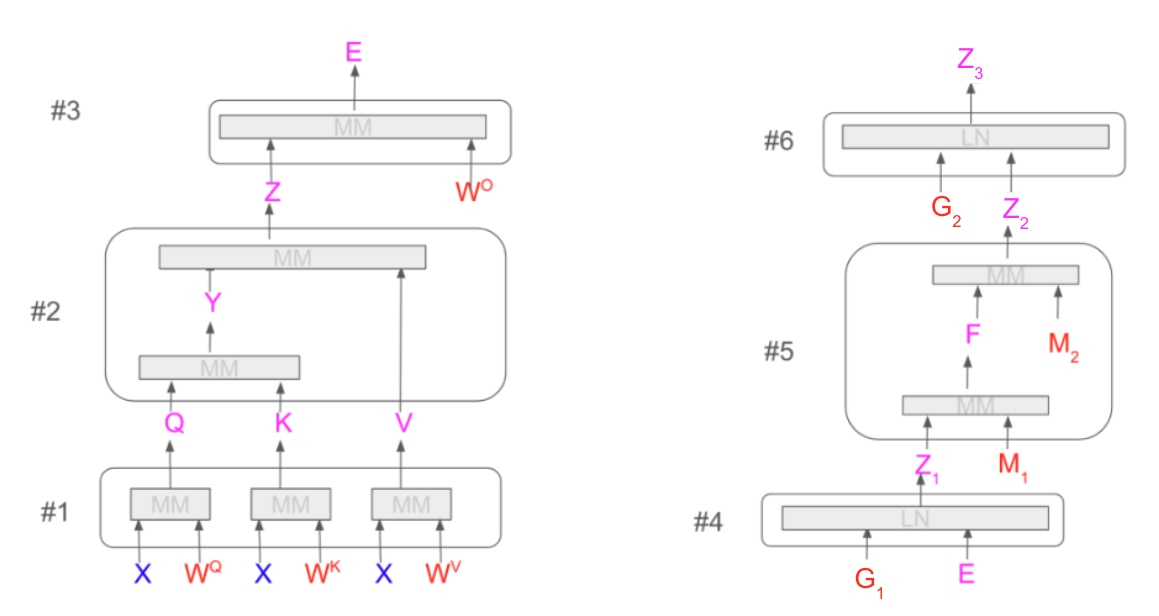}
    \caption{A diagram of the Transformer block architecture.}
    \label{fig:transformer_block}
\end{figure}

In tensor-level MoR, a single quantization type is selected for an entire tensor, though different
tensors within a model can be quantized to different types. Take Figure~\ref{fig:transformer_block} as an example, 
layer \#1 is the linear projection module, layer \#2 is the attention module, layer
\#3 is the out projection module, layer \#4 is the layer norm module, layer \#5 is the 
MLP module, and layer \#6 is the layer norm module. Suppose we want to quantize
the MLP module \#5, the tensors could be treated differently: 
the weight tensors $M_1$, $M_2$, and the activation tensor $Z_1$ might be quantized to E4M3, while the
activation tensor $F$ remains in BF16.

We implemented and evaluated a specific tensor-level MoR recipe based on our framework
(Algorithm~\ref{algo:mor_framework}). In this recipe, the ordered list of representation types is
[E4M3, BF16]. The acceptance metric for E4M3 is that the tensor's mean relative quantization
error, calculated over all non-zero elements, must be less than a threshold $th_{E4M3}$:
\begin{align}
n \leftarrow \sum_{x_{i, j} \in X, x_{i, j} \neq 0} 1 \\  
\text{error} = \frac{1}{n} \sum_{x_{i, j} \in X, x_{i, j} \neq 0} \left| \frac{x_{i,j} - Q(x_{i,j})}{x_{i,j}} \right| < th_{E4M3}
\label{eq:tensor_relative_error}
\end{align}
where $n$ is the count of non-zero elements in tensor $X$, and $Q$ is the function that quantizes
an element from its original format (e.g., BF16) to E4M3. If the computed error exceeds the
threshold, the E4M3 quantization is rejected, and the entire tensor reverts to BF16.

\begin{figure}[htbp]
    \centering
    \includegraphics[width=\textwidth]{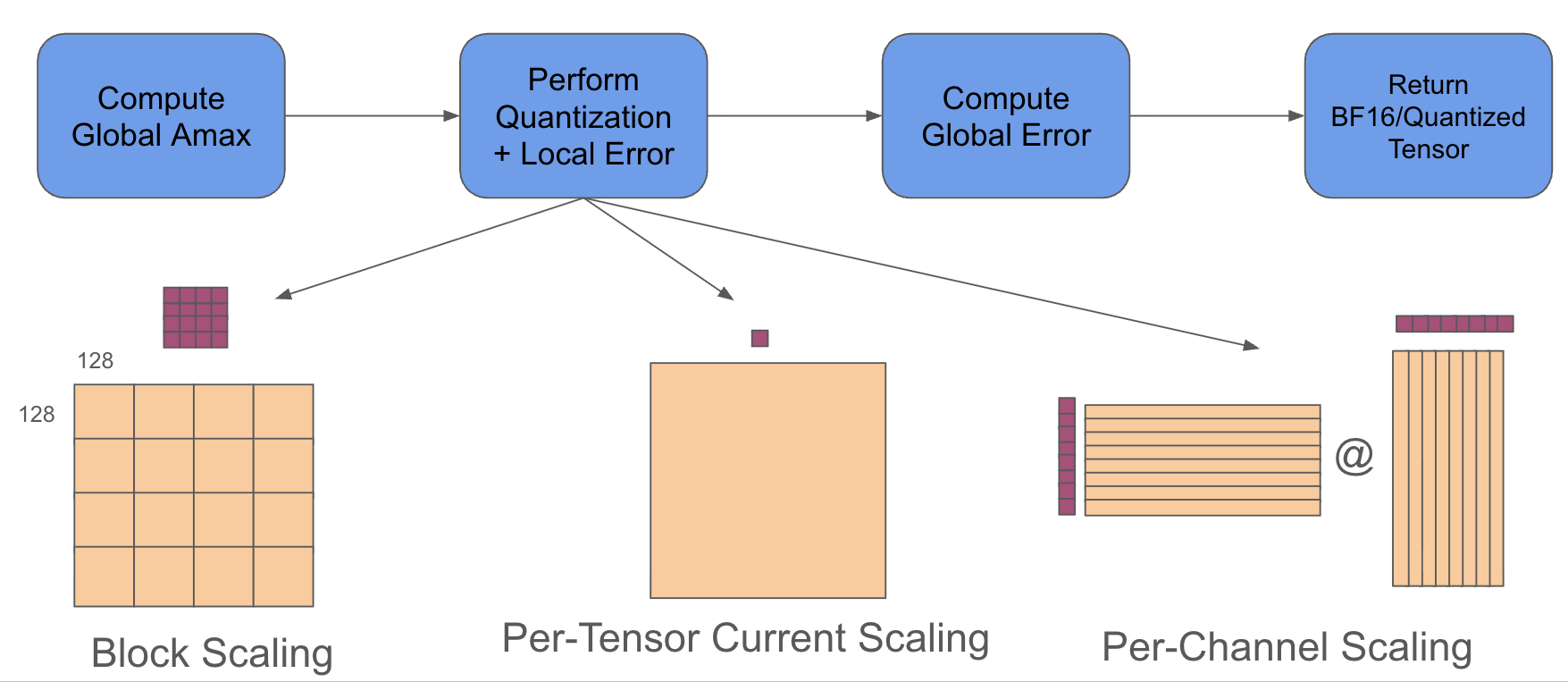}
    \caption{Illustration of tensor-level MoR. A single decision (E4M3 or BF16) is made for the
    entire tensor $X$. This decision is based on a global relative error, which can be computed using
    various internal partitioning strategies for the quantization step (e.g., per-block, no-partition, per-channel).
    The process uses the GAM scaling algorithm from Algorithm~\ref{algo:gam_scaling} to determine a
    shared mantissa and per-partition exponents (purple squares).}
    \label{fig:mor_tensor_alg}
\end{figure}

A key aspect of this approach is that while the final decision (E4M3 or BF16) is global to the tensor,
the underlying quantization and error calculation can leverage different partitioning strategies. 
Figure \ref{fig:mor_tensor_alg}
illustrates this idea. We first computes the global absolute maximum value to
retrieve the global mantissa $M$ using the GAM scaling algorithm (Algorithm~\ref{algo:gam_scaling}).
Then we quantize the tensor into E4M3 with different partition strategies.
We partition the tensor into $128\times 128$ blocks, keep the full tensor
non-partitioned, or partition the tensor based on the dot product channel. 
The purple squares in the figure represent the E8M0 scaling factors that are required for
different partitioning strategies. We compute the local quantization error
using the relative error metric for each block $b$ based on the partitioning strategy, 
and then aggregate the local errors into the global quantization error. This global error is
then compared against the threshold to make the final, tensor-wide decision.
The primary advantage of this approach is to isolate model quality 
considerations from representation efficiency considerations.
The hypothesis is in the following: as long as the relative error of the
representation is bounded, we can guarantee the model quality to be on par compared to
the BF16 baseline. So we are free to explore different partitioning strategies
to further improve the representation efficiency as long as the relative
error bound condition is met.

\subsection{MoR at SubTensor Granularity}\label{sec:subtensor}

\begin{figure}[htbp]
    \centering
    \includegraphics[width=\textwidth]{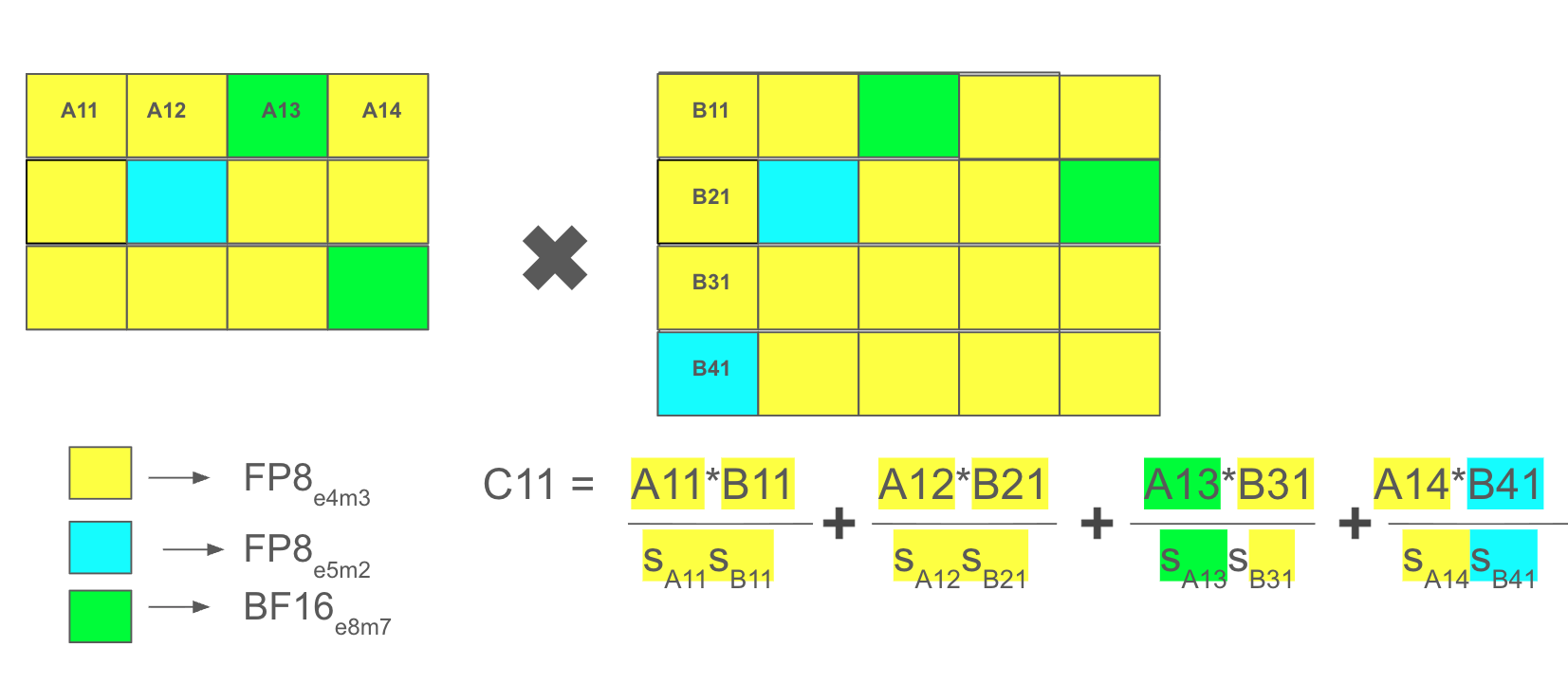}
    \caption{An illustration of sub-tensor MoR. Blocks within matrices A and B are quantized to
    different formats (E4M3, E5M2, BF16). The corresponding GEMM operation may require upcasting
    blocks (e.g., $B_{31}$) to a higher precision (BF16) before multiplication.}
    \label{fig:mor_subtensor}
\end{figure}

MoR at sub-tensor granularity means we can apply different quantization types to different
partition blocks. Figure \ref{fig:mor_subtensor} illustrates such concept. The 
yellow blocks are quantized to the E4M3 type; the blue blocks are quantized to
E5M2 type; and the green blocks are not quantized and stay at the original
BF16 type (assuming the original input tensor type is BF16). The GEMM 
operation is much more complicated in this case, as we might need to 
perform GEMM on blocks with different types. If there is no hardware 
support to perform the dot product of two different tensor types, 
we will upcast the quantized block to match the operand with higher precision.
For example, in Figure \ref{fig:mor_subtensor}, $A_{13}$ is in BF16, while $B_{31}$ is in E4M3,
if there is no hardware support for such dot product, we will upcast $B_{31}$ to BF16 and
perform BF16 GEMM between the two blocks.

Sub-tensor MoR applies the framework's decision logic at the block level, allowing different blocks
within the same tensor to have different quantization types. This concept is illustrated in
Figure~\ref{fig:mor_subtensor}, where a single tensor is composed of blocks in E4M3 (yellow),
E5M2 (blue), and the original BF16 format (green). The GEMM 
operation is much more complicated in this case, as we might need to 
perform GEMM on blocks with different types. If there is no hardware 
support to perform the dot product of two different tensor types, 
we will upcast the quantized block to match the operand with higher precision.
For example, in Figure \ref{fig:mor_subtensor}, $A_{13}$ is in BF16, while $B_{31}$ is in E4M3,
if there is no hardware support for such dot product, we will upcast $B_{31}$ to BF16 and
perform BF16 GEMM between the two blocks.

We developed and evaluated two sub-tensor MoR recipes based on our framework (Algorithm~\ref{algo:mor_framework}).
\paragraph{Algorithm 1: Three-Way Selection (E4M3/E5M2/BF16)}
The first algorithm uses an ordered list of types [E4M3, E5M2, BF16]. The decision process for a block $b$ is as follows:
\begin{enumerate}
    \item \textbf{Check E4M3:} The block is quantized to E4M3 if its total relative error is lower than
    if it were quantized to E5M2. This is the metric $M_1$:
    \begin{align}
        \sum_{x \in b, x \neq 0} \left| \frac{x - Q_{E4M3}(x)}{x} \right| < \sum_{x \in b, x \neq 0} \left| \frac{x - Q_{E5M2}(x)}{x} \right|
        \label{eq:subtensor_relative}
    \end{align}
    \item \textbf{Check E5M2:} If the E4M3 check fails, the framework then checks if the block's dynamic range
    fits within the representable normal range of E5M2 (the maximum value representable by the E5M2 type is 57344, and the 
smallest value in the normal range of the E5M2 type is $2^{-14}$). This is the metric $M_2$:
    \begin{align}
        \frac{\max(\text{abs}(b))}{\min(\text{abs}(b))} < \frac{57344}{2^{-14}}
        \label{eq:subtensor_range}
    \end{align}
    \item \textbf{Fallback:} If both checks fail, the block remains in BF16.
\end{enumerate}

\paragraph{Algorithm 2: Two-Way Selection (E4M3/BF16)}
The second algorithm is a simpler variation that uses the ordered list [E4M3, BF16].
It uses the exact same metric as in Equation~\ref{eq:subtensor_relative} to evaluate E4M3. However, the logic
is different: if E4M3 quantization yields a higher relative error than E5M2, the framework does not
consider E5M2 as an alternative. Instead, it immediately falls back to BF16. In this recipe, E5M2 serves
only as a high-quality benchmark to determine if E4M3 is a suitable choice, but it is never selected as the
final format.
\section{Experiments}\label{sec:exps}

All experiments were performed on the Nemotron-3 8B model \cite{nemotron-8b}, a dense transformer model
with 32 transformer blocks.
Our quantization strategy targets only the linear layers within each transformer block.
Specifically, we apply the MoR algorithms
to four linear layers in one transformer block. For each linear layer, we apply MoR on 
the activation, weight, and gradient tensors and their transposes for the forward and
backward pass GEMM operations.
Using Figure~\ref{fig:transformer_block} 
as an illustration, we apply MoR on layer \#1: Linear QKV, this layer projects the 
input tensor to the Q, K, and V space; layer \#3, Linear Projection, this layer projects the
attention output to an output dimension; layer \#5, FC1 and FC2, these are the two fully connected
layers in the MLP module. All models were trained using the
Megatron-LM framework \cite{megatronlm}.

To evaluate the robustness of our MoR algorithms, we trained the model using two distinct configurations,
varying both the dataset and key hyperparameters. In the first configuration, the model was trained for one
trillion tokens on a sample from the Nemotron-4 training data \cite{nemotron4paper}, with a cosine
learning rate schedule annealing from $3 \times 10^{-4}$ to $3 \times 10^{-5}$ and a batch size of 1024.
The second configuration used one trillion tokens from the higher-quality Nemotron-H dataset
\cite{nemotronhpaper}, a cosine learning rate schedule annealing from $1.2 \times 10^{-3}$ to $3 \times 10^{-6}$, and a
batch size of 1536. These details are summarized in Table~\ref{tab:train_configs}. The second configuration
consistently achieved lower training loss and superior scores on downstream tasks.

\begin{table}[htbp]
    \centering
    \caption{A comparison of the two training configurations.}
    \label{tab:train_configs}
    \begin{tabular}{l l l}
        \toprule
        \textbf{Parameter} & \textbf{Configuration 1} & \textbf{Configuration 2} \\
        \midrule
        Training Data      & Nemotron4 \cite{nemotron4paper} & NemotronH \cite{nemotronhpaper} \\
        Training Tokens    & 1 Trillion               & 1 Trillion \\
        LR Schedule        & Cosine                   & Cosine \\
        Peak Learning Rate & $3 \times 10^{-4}$       & $1.2 \times 10^{-3}$ \\
        Final Learning Rate& $3 \times 10^{-5}$       & $3 \times 10^{-6}$ \\
        Batch Size         & 1024                     & 1536 \\
        \bottomrule
    \end{tabular}
\end{table}

\begin{figure}[htbp]
    \centering
    \includegraphics[width=\textwidth]{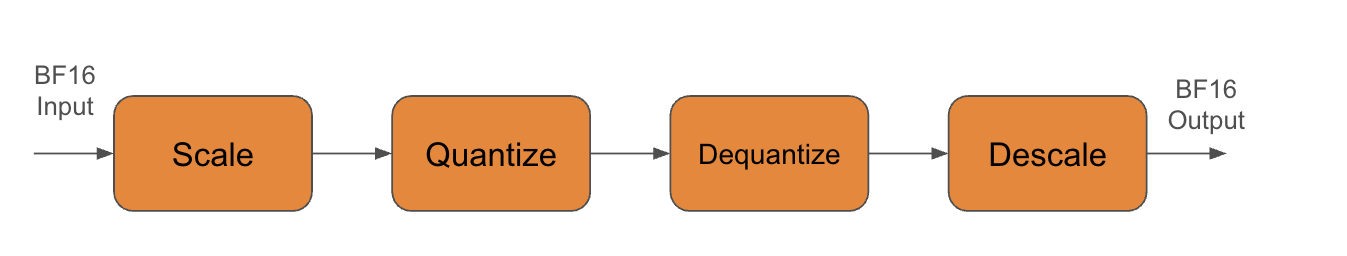}
    \caption{The fake quantization workflow. The input and output data type are kept in BF16 precision,
    but the process simulates the precision loss of the target format.}
    \label{fig:fake_quantize}
\end{figure}

To simulate the effects of quantization while maintaining a standard training pipeline, we employed a fake
quantization framework, illustrated in Figure~\ref{fig:fake_quantize}. In this process, a BF16 input block is
first scaled and quantized to the target format (e.g., E4M3). It is then immediately dequantized back to BF16
and de-scaled to its original numerical range. The output tensor therefore remains in the BF16 format, but it
carries the information loss characteristic of the lower-precision representation.

In the following sections, we present a comprehensive analysis of the tensor-level MoR algorithm, including its
impact on model quality across multiple block dimensions and a detailed analysis of its decision-making process.
Our evaluation of the sub-tensor MoR algorithm is more concise, focusing primarily on its effect on final
model quality.
\subsection{Experiments for MoR at Tensor Granularity}\label{sec:exp_tensor}

\subsubsection{MoR with Different Partition Strategies}

To study how MoR behaves across different scaling granularities, we implemented and evaluated three distinct
partition strategies for the tensor-level MoR algorithm:
\begin{itemize}
    \item \textbf{Per-Block:} Computes one scaling factor for each $128 \times 128$ block of the tensor.
    \item \textbf{Per-Tensor:} Computes a single scaling factor for the entire tensor.
    \item \textbf{Per-Channel:} Computes a scaling factor for each row of the first GEMM operand and each column of the second, aligned with the dot-product dimension.
\end{itemize}
To ensure a fair comparison, we used the same acceptance threshold, $th_{E4M3}$, for all three strategies.

To select an appropriate threshold, we analyzed the activation, weight, and gradient tensors from a 
reference BF16 training run, collected near the end of training. 
We calculated the relative error that would result from quantizing these tensors to E4M3. 
Based on this analysis, we determined that a threshold of $th_{E4M3} = 4.5\%$ would quantize 
approximately 95\% of the tensors to E4M3, while leaving the remaining 5\% in BF16. 
This ratio provided a reasonable starting point for our experiments, 
so we used a fixed threshold of 4.5\% for all tensor-level MoR variants.

\begin{figure}[htbp]
    \centering 
    
    \subfloat[Training Loss.]{\includegraphics[width=1.0\textwidth]{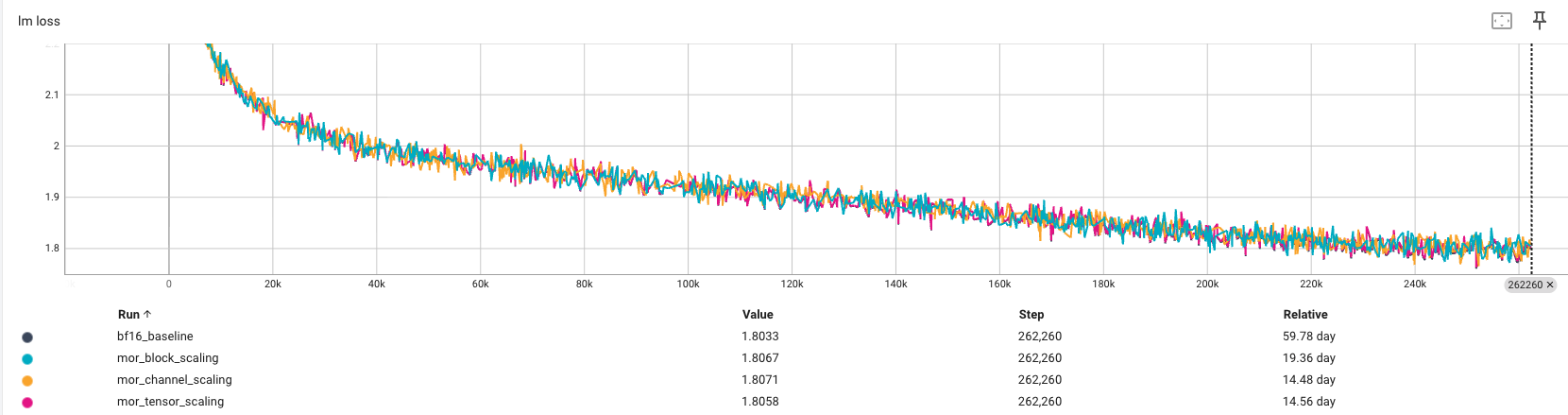}}
    
    \subfloat[Validation Loss.]{\includegraphics[width=1.0\textwidth]{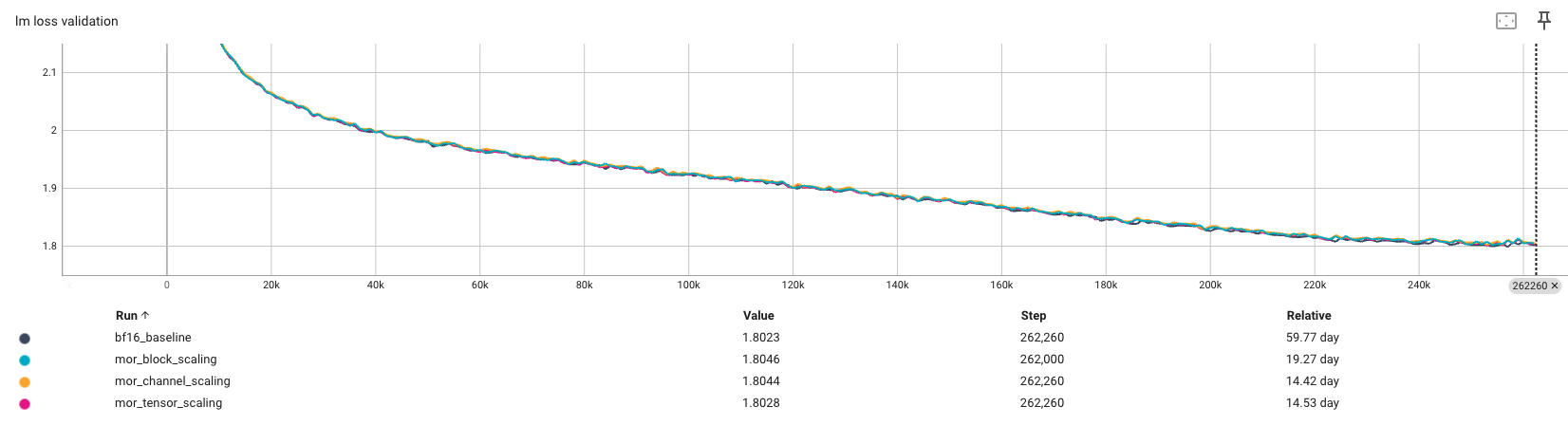}}
    
    \subfloat[L2 Norm of the Parameters.]{\includegraphics[width=1.0\textwidth]{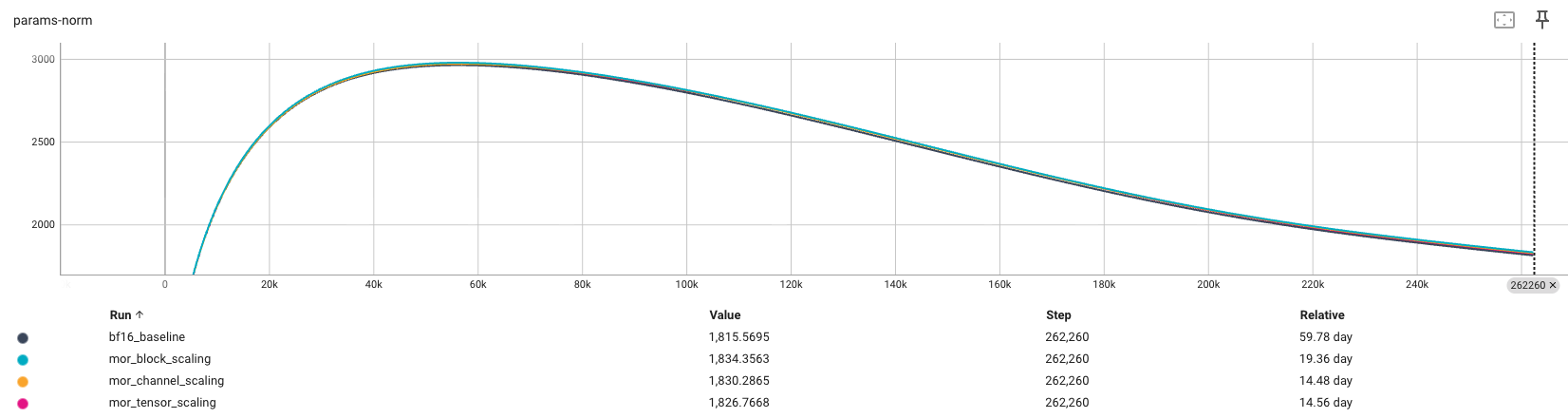}}
    
    \caption{Training loss, validation loss, and the parameter L2 Norm using the first training configuration. In the legend, from top to bottom, the curves represent: BF16 Baseline, MoR Per-Block, MoR Per-Channel, and MoR Per-Tensor.}
    \label{fig:train_nemotron4}
\end{figure}

\begin{figure}[htbp]
    \centering 
    
    \subfloat[Training Loss.]{\includegraphics[width=1.0\textwidth]{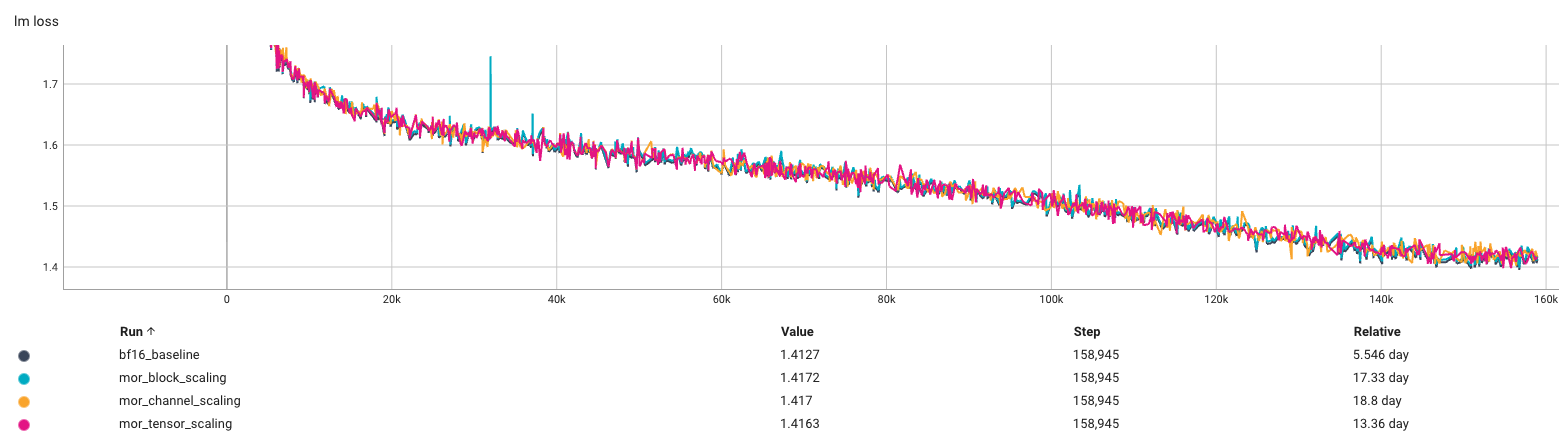}}
    
    \subfloat[Validation Loss.]{\includegraphics[width=1.0\textwidth]{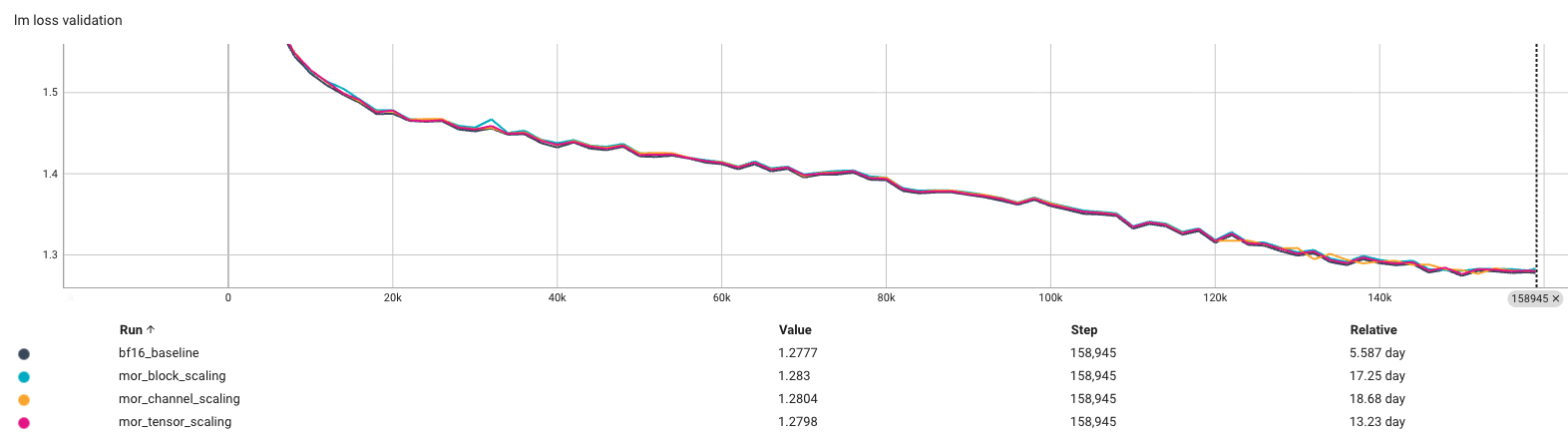}}
    
    \subfloat[L2 Norm of the Parameters.]{\includegraphics[width=1.0\textwidth]{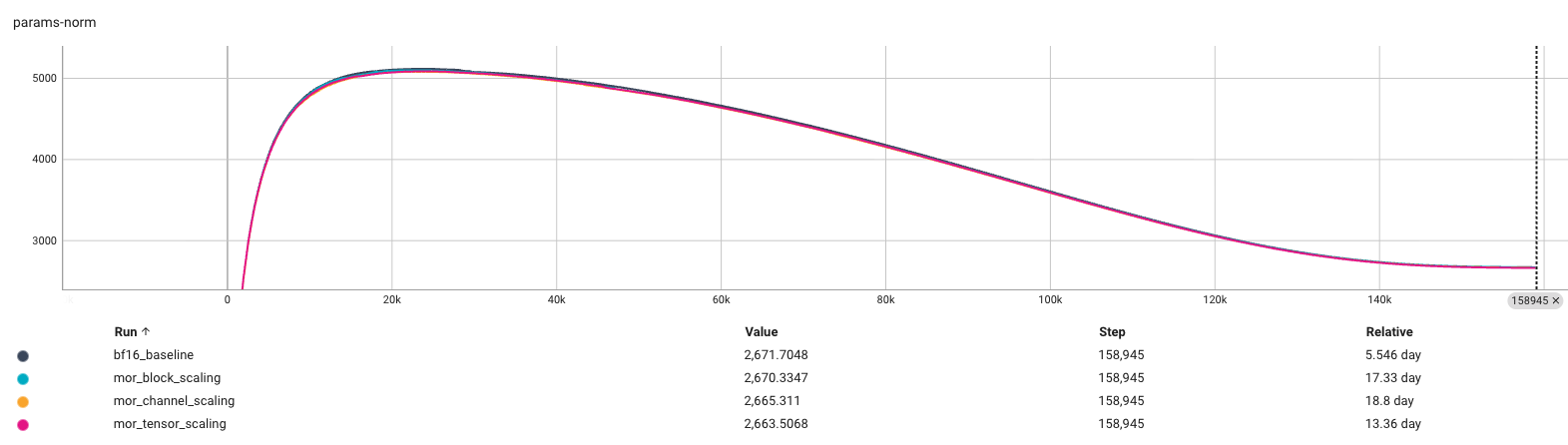}}
    
    \caption{Training loss, validation loss, and the parameter L2 Norm using the second training configuration. In the legend, from top to bottom, the curves represent: BF16 Baseline, MoR Per-Block, MoR Per-Channel, and MoR Per-Tensor.}
    \label{fig:train_nemotron5}
\end{figure}

The training dynamics for both training configurations are shown in 
Figures~\ref{fig:train_nemotron4} and \ref{fig:train_nemotron5}. Across both configurations, 
the training loss, validation loss, and parameter norm curves for all 
MoR variants closely track the BF16 baseline, indicating stable training process.

Table~\ref{tab:model_performance} summarizes the final loss values and downstream task evaluation scores
using the last saved checkpoint. The final training and validation losses for all MoR variants are within 
0.5\% of the BF16 baseline. While the downstream task evaluations show higher variance, the three MoR strategies 
generally achieve on-par results with the baseline (most differences are about 1\%). 
Some benchmarks, such as OpenBookQA and CommonSenseQA under Configuration 1, show a larger deviation. 
We attribute this to checkpoint selection sensitivity; selecting for the best score across several recent checkpoints, 
rather than just the last one, closes this gap significantly. For example, the best OpenBookQA 
scores under Configuration 1 are 42.2 (Block), 42.2 (Tensor), and 42.8 (Channel), which are competitive with the 42.8 baseline;
similarly, the best CommonSenseQA under Configuration 1 are 37.84 (Block), 32.76 (Tensor), and 34.32 (Channel), which
are closer to the 34.32 baseline.

\begin{figure}[htbp]
    \centering
    
    \subfloat[Training Configuration 1.]{\includegraphics[width=0.7\textwidth]{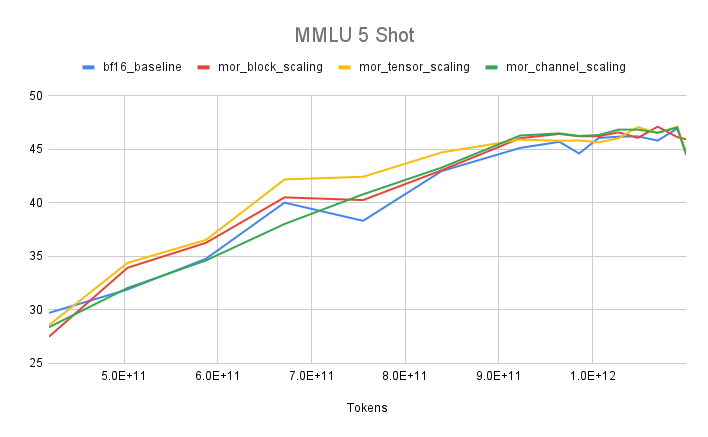}}
   
    \subfloat[Training Configuration 2.]{\includegraphics[width=0.7\textwidth]{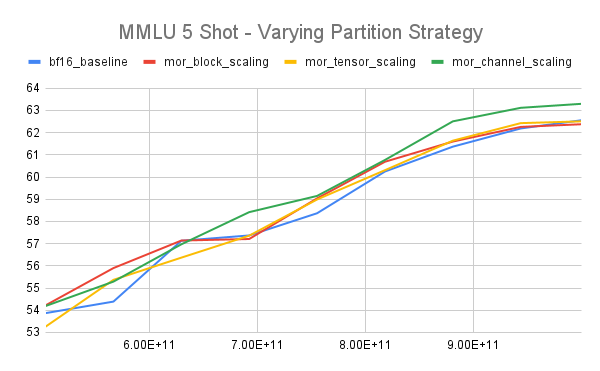}}
    
    \caption{The MMLU 5-shot scores over the training progress for the two training configurations.}
    \label{fig:mmlu5shot}
\end{figure}

This trend of strong performance is further supported by the MMLU 5-shot evaluation progress, 
shown in Figure~\ref{fig:mmlu5shot}. The MoR curves track the BF16 baseline closely, 
with the per-channel MoR variant slightly outperforming the baseline in Configuration 2. 
These observations lead us to conclude that the tensor-level MoR algorithm is robust 
across different partitioning strategies and training configurations, 
consistently achieving model quality on-par with the BF16 baseline.

\begin{table}[htbp]
    \centering
    \caption{Model quality comparison with varying partition strategies.}
    \label{tab:model_performance}
    \begin{tabular}{l cccc cccc}
        \toprule
        & \multicolumn{4}{c}{\textbf{Configuration 1}} & \multicolumn{4}{c}{\textbf{Configuration 2}} \\
        \cmidrule(lr){2-5} \cmidrule(lr){6-9}
        \textbf{Metric} & \textbf{BF16} & \textbf{Block} & \textbf{Tensor} & \textbf{Channel} & \textbf{BF16} & \textbf{Block} & \textbf{Tensor} & \textbf{Channel} \\
        \midrule
        Training Loss    & \textbf{1.8033} & 1.8067 & 1.8058 & 1.8071 & \textbf{1.4127} & 1.4172 & 1.4163 & 1.4170 \\
        Validation Loss  & \textbf{1.8023} & 1.8046 & 1.8028 & 1.8044 & \textbf{1.2777} & 1.2830 & 1.2798 & 1.2804 \\
        \midrule
        MMLU             & 44.72  & \textbf{45.95}  & 44.61  & 44.50  & 62.56  & 62.38  & 62.51  & \textbf{63.30}  \\
        WinoGrande       & 66.69  & 66.77  & 66.69  & \textbf{67.40}  & 69.85  & \textbf{72.38}  & 71.67  & 70.64  \\
        PIQA             & 78.45  & \textbf{79.16}  & 77.69  & 78.40  & \textbf{81.61}  & 80.63  & 81.07  & 80.96  \\
        HellaSwag        & 74.93  & 74.84  & 74.85  & \textbf{75.02}  & 77.56  & 77.47  & \textbf{77.82}  & 77.67  \\
        Arc-Easy         & 73.48  & 73.36  & 73.91  & \textbf{74.24}  & 83.67  & \textbf{83.88}  & 82.41  & 81.82  \\
        Arc-Challenge    & 41.30  & \textbf{43.34}  & 42.24  & 42.24  & 54.78  & \textbf{58.19}  & 56.48  & 56.23  \\
        OpenBookQA       & \textbf{42.80}  & 39.80  & 40.40  & 41.80  & 44.00  & \textbf{45.20}  & 44.80  & \textbf{45.20}  \\
        SIQA             & 44.63  & 45.65  & \textbf{46.42}  & 45.04  & \textbf{46.83}  & 45.45  & 46.72  & 46.57  \\
        CommonSenseQA    & \textbf{34.32}  & 34.23  & 30.88  & 26.37  & \textbf{67.65}  & 65.93  & 66.34  & 64.78  \\
        \bottomrule
    \end{tabular}
\end{table}

\subsubsection{Ablation Study on MoR Settings}

In this section, we study how different MoR settings affect model quality. We performed three sets of
ablation experiments using the per-block partitioning strategy and the first training configuration.

The first experiment varies the \textbf{block dimension}. By reducing the block size from $128 \times 128$ to
$64 \times 64$, we hypothesize that the more granular scaling factors will lead to a more accurate
quantized representation. The second experiment adjusts the \textbf{acceptance threshold}. By increasing the
threshold from 4.5\% to 5.0\%, we allow more tensors with higher relative error to be quantized to E4M3,
potentially impacting model quality. The third experiment compares different \textbf{scaling algorithms}: our
proposed GAM scaling, a simpler E8M0 per-block scaling, and the standard per-block FP32 amax scaling.

\begin{figure}[htbp]
    \centering 
    
    \subfloat[Training Loss.]{\includegraphics[width=1.0\textwidth]{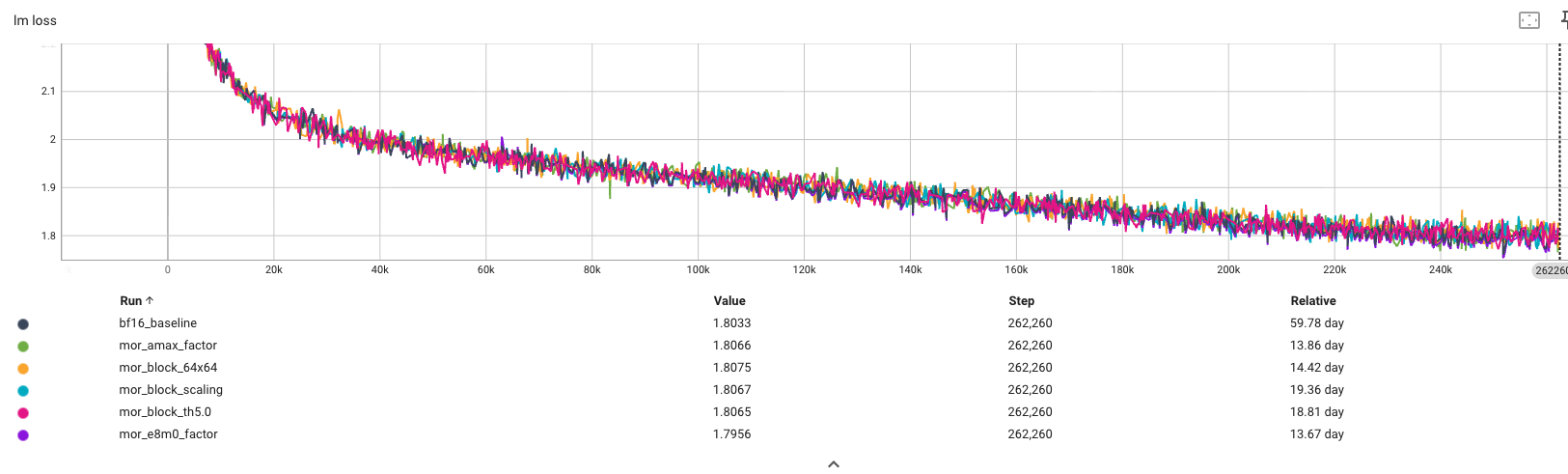}}
    
    \subfloat[Validation Loss.]{\includegraphics[width=1.0\textwidth]{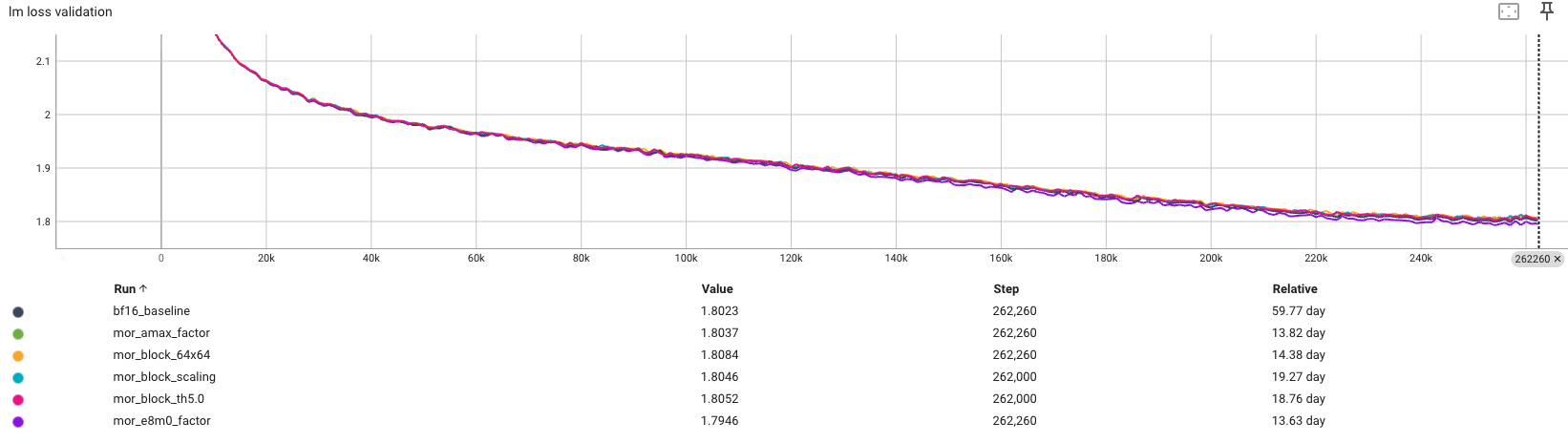}}
    
    \subfloat[L2 Norm of the Parameters.]{\includegraphics[width=1.0\textwidth]{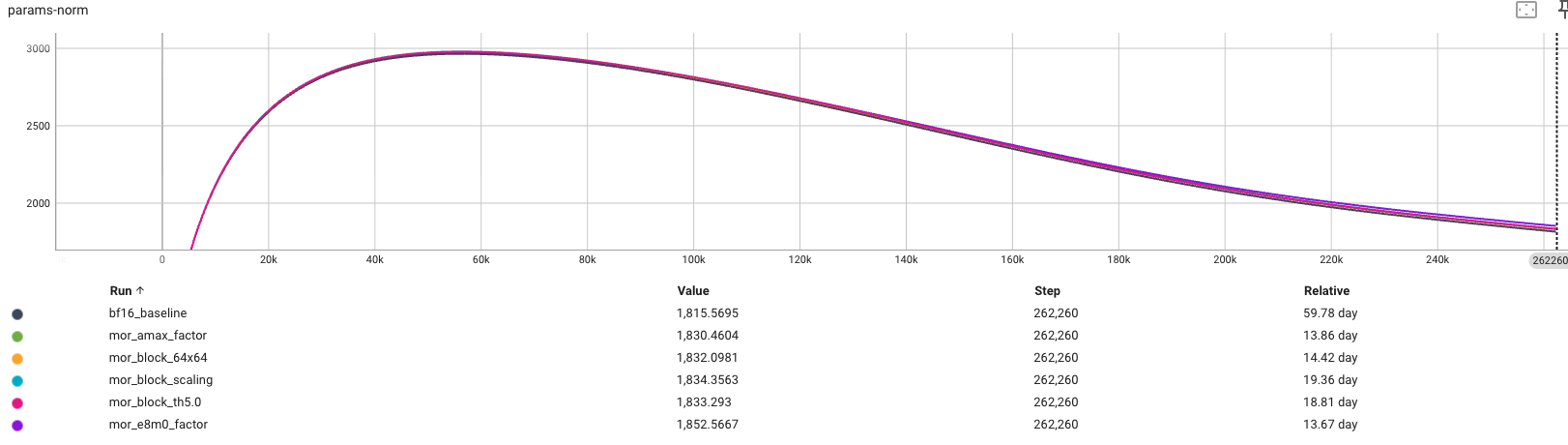}}
    
    \caption{Training loss, validation loss, and the parameter L2 Norm using the first training configuration. In the legend, from top to bottom, the curves represent: BF16 Baseline, MoR FP32 Amax, MoR 64x64 Block, MoR 128x128 Block (Default), MoR 5.0\% Threshold, and MoR E8M0.}
    \label{fig:train_compare}
\end{figure}

The training dynamics for these experiments are summarized in Figure~\ref{fig:train_compare}. All variants
track the BF16 baseline closely, indicating stable training. Notably, the E8M0 scaling variant consistently
achieved a slightly lower training and validation loss than the BF16 baseline. The final metrics and
downstream evaluation scores are presented in Table~\ref{tab:eval_morset}. While final loss values are
all within 0.5\% of the baseline, the downstream task results show more variance. The standard FP32 amax
scaling strategy, for instance, results in an MMLU score of 1.63\% lower than the baseline. Conversely, some
high-variance benchmarks like CommonSenseQA show unexpectedly large gains for the $64 \times 64$ and 5.0\%
threshold settings, which may be attributable to evaluation noise.

\begin{figure}[htbp]
    \centering
    
    \subfloat[Varying block size and E4M3 threshold.]{\includegraphics[width=0.7\textwidth]{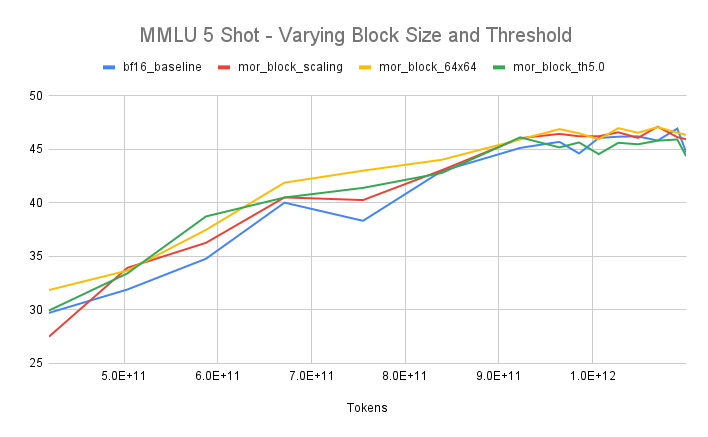}}
   
    \subfloat[Varying scaling factor algorithm.]{\includegraphics[width=0.7\textwidth]{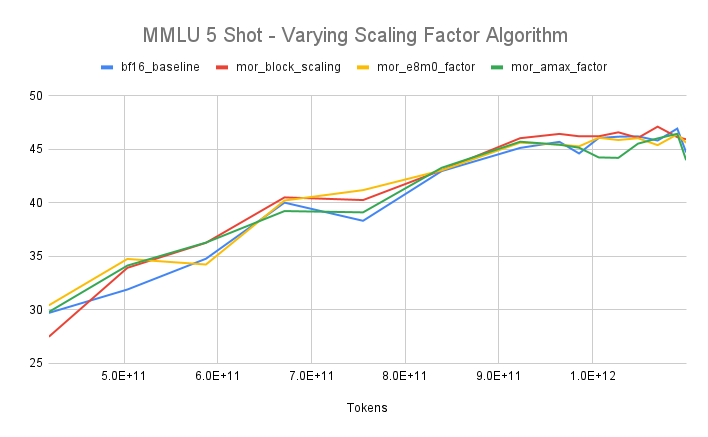}}
    
    \caption{The MMLU 5-shot scores over the training progress when varying the MoR block size, the MoR E4M3 threshold, and the scaling factor algorithms.}
    \label{fig:mmlu5shot_morset}
\end{figure}

The MMLU performance over the course of training, shown in Figure~\ref{fig:mmlu5shot_morset}, provides
clearer insights. As expected, the smaller $64 \times 64$ block size closely tracks the baseline and the
default $128 \times 128$ MoR performance, confirming the benefit of finer granularity. Increasing the
threshold to $5.0\%$ (quantizing more tensors to E4M3) results in a slight degradation in MMLU score in the
later stages of training. The comparison of scaling algorithms in Figure~\ref{fig:mmlu5shot_morset}(b) is
particularly revealing. The E8M0 scaling factor tracks the BF16 baseline most closely, while the standard
FP32 amax scaling creates a noticeable gap. This suggests that having a consistent mantissa
across the tensor, a feature of both our GAM algorithm and the E8M0 format (which has no mantissa bits),
plays a crucial role in maintaining stable training dynamics and final model performance.

\begin{table}[htbp]
    \centering
    \caption{Model quality comparison with varying MoR settings.}
    \label{tab:eval_morset}
    \begin{tabular}{l cccccc}
        \toprule
        \textbf{Metric} & \textbf{BF16} & \textbf{Block 128x128} & \textbf{Block 64x64} & \textbf{Th5.0\%} & \textbf{Amax Factor} & \textbf{E8M0 Factor} \\
        \midrule
        Training Loss    & 1.8033 & 1.8067 & 1.8075 & 1.8065 & 1.8066 & \textbf{1.7956} \\
        Validation Loss  & 1.8023 & 1.8046 & 1.8084 & 1.8052 & 1.8037 & \textbf{1.7946} \\
        \midrule
        MMLU             & 44.72  & 45.95  & \textbf{46.37}  & 44.37  & 43.99  & 45.62  \\
        WinoGrande       & 66.69  & 66.77  & 67.17  & 67.56  & \textbf{68.75}  & 67.48  \\
        PIQA             & 78.45  & \textbf{79.16}  & 77.86  & 78.18  & 77.48  & 79.11  \\
        HellaSwag        & 74.93  & 74.84  & 75.08  & 75.06  & \textbf{75.13}  & 74.69  \\
        Arc-Easy         & 73.48  & 73.36  & 73.27  & \textbf{73.70}  & 73.06  & 72.60  \\
        Arc-Challenge    & 41.30  & \textbf{43.34}  & 42.32  & 42.24  & 42.06  & 42.75  \\
        OpenBookQA       & \textbf{42.80}  & 39.80  & 41.80  & 41.40  & 39.40  & 40.40  \\
        SIQA             & 44.63  & \textbf{45.65}  & \textbf{45.65}  & 45.04  & 44.83  & 44.47  \\
        CommonSenseQA    & 34.32  & 34.23  & 39.15  & \textbf{39.39}  & 33.74  & 36.61  \\
        \bottomrule
    \end{tabular}
\end{table}

\subsubsection{Tensor Statistics}

The MoR framework's reliance on relative error allows us to collect detailed statistics for each tensor on a per-mini-batch basis. 
This section presents these statistics, captured over the course of training, to provide deeper 
insights into the algorithm's dynamic decision-making process.

\begin{figure}[htbp]
    \centering
    \includegraphics[width=0.7\textwidth]{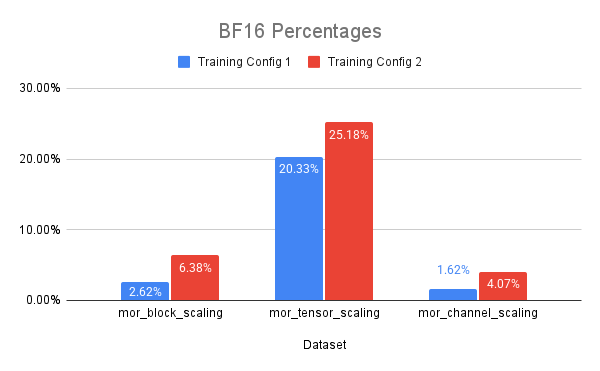}
    \caption{Percentage of tensors that fall back to the BF16 type.}
    \label{fig:bf16_perc}
\end{figure}

The overall percentage of tensors that fall back to BF16 throughout training is summarized in Figure~\ref{fig:bf16_perc}. 
The second training configuration, which uses higher-quality data, consistently requires more tensors to remain in BF16. 
Across the three partitioning strategies, per-channel scaling exhibits the highest quantization efficiency, 
with only $1.62\%$ and $4.07\%$ of tensors falling back to BF16 for the first and second configurations, respectively. 
Conversely, the per-tensor strategy is the least efficient, requiring the most fallbacks.

\begin{figure}[htbp]
    \centering
    \includegraphics[width=0.7\textwidth]{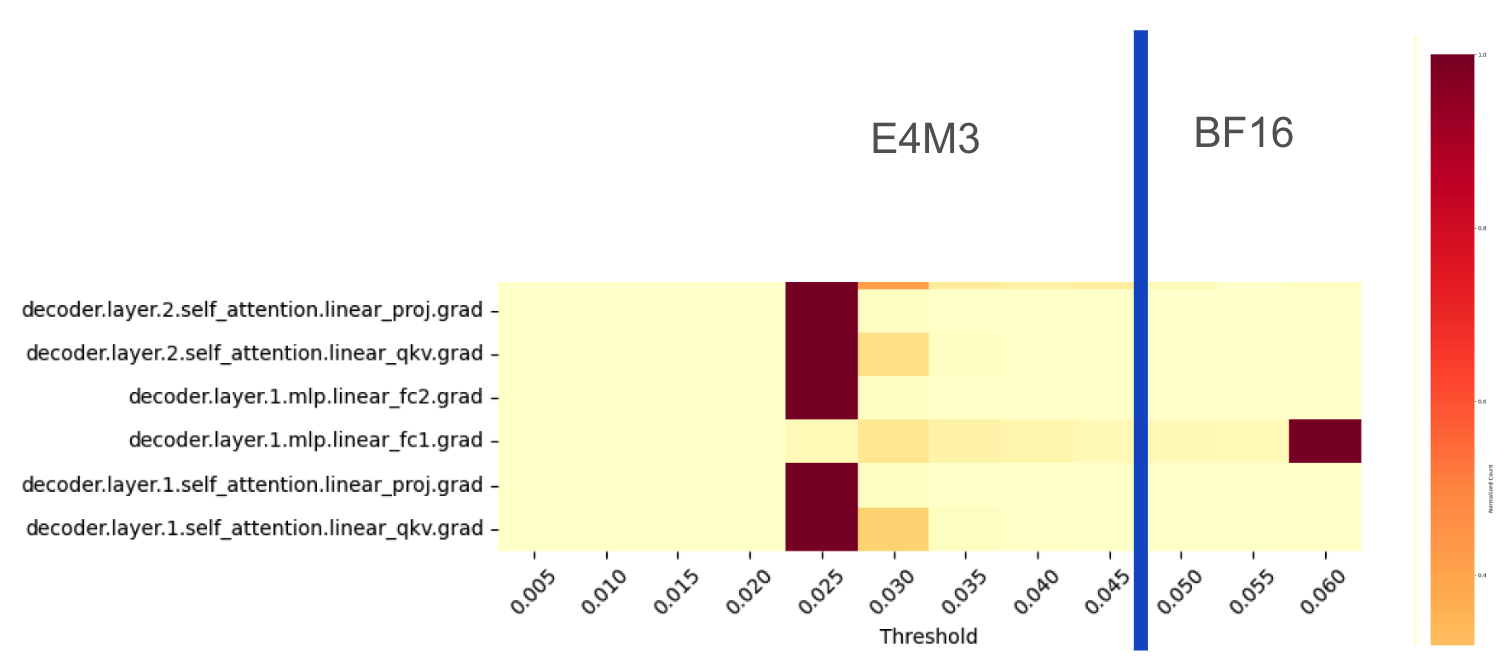}
    \caption{Annotation for the relative error histogram visualization. The x axis is the relative error histogram bins. The y axis is the name of the tensor.}
    \label{fig:annotation}
\end{figure}

To visualize the distribution of relative error, we generate periodic histograms 
for each tensor, as annotated in Figure~\ref{fig:annotation}. The \textbf{x-axis} of our 
heatmaps represents the bins of these histograms, where each bin covers a $0.5\%$ range of relative error. 
The first bin corresponds to the normalized counts for relative error < $0.5\%$, 
and second bin corresponds to the normalized counts for $0.5\% \leq$  relative error < $1.0\%$, and so forth. 
The right most bin corresponds to relative error > $5.5\%$ and beyond. 
The blue vertical line marks the E4M3 threshold $4.5\%$. 
So tensors to the left of the line will be quantized to E4M3, 
and tensors to the right of the line will fall back to BF16.
The \textbf{y-axis} in the heatmap is the name of the tensor. 
"decoder.layer.n" means this is the nth transformer block.
"mlp" vs "self\_attention" means this is the MLP module or the attention module.
"linear\_qkv", "linear\_proj", "fc1", "fc2" are the linear modules in the attention/MLP modules.
"grad", "weight", "input/ln\_out" are the gradient, weight, and activation tensors 
accessed in the linear module.

We also collected finer-grained statistics to learn the distribution of the relative
error. The statistics and the visualization can be illustrated in Figure \ref{fig:annotation}.
When we compute the relative error of one tensor, we will get one count, and we can
accumulate the count to a histogram. The x axis in the heatmap visualization is the relative 
error histogram.
Each bin covers the relative error value in a $0.5\%$ range. 
The first bin corresponds to the normalized counts for relative error < $0.5\%$, 
and second bin corresponds to the normalized counts for $0.5\% \leq$  relative error < $1.0\%$, and so forth. 
The right most bin corresponds to relative error > $5.5\%$ and beyond. 
The blue vertical line marks the E4M3 threshold $4.5\%$. 
So tensors to the left of the line will be quantized to E4M3, 
and tensors to the right of the line will fall back to BF16.
The y axis in the heatmap is the name of the tensor. 
"decoder.layer.n" means this is the nth transformer block.
"mlp" vs "self\_attention" means this is the MLP module or the attention module.
"linear\_qkv", "linear\_proj", "fc1", "fc2" are the linear modules in the attention/MLP modules.
"grad", "weight", "input/ln\_out" are the gradient, weight, and activation tensors 
accessed in the linear module. 

One mini-batch contributes to one count in the histogram.
So the sum over each row is the number of mini-batches that are processed.
We then normalize the counts over each row into a number between 0 and 1.
The normalized counts are visualized with different heat colors, 
with darker color implying denser counts in the histogram bin, and 
lighter color implying sparser counts in the histogram bin.
In order to see how the distribution changes over time, we reset 
the histogram every 6000 steps.

There are 32 transformer blocks in the model, each block has 4 linear layers, and each layer has 3 tensors.
In total there are $32 \times 4 \times 3 = 384$ rows in the heatmap. The full heatmaps are too large to display.
We therefore present representative subsets in the following analyses.

\begin{figure}[htbp]
    \centering
    
    \subfloat[First three transformer blocks.]{\includegraphics[width=0.7\textwidth]{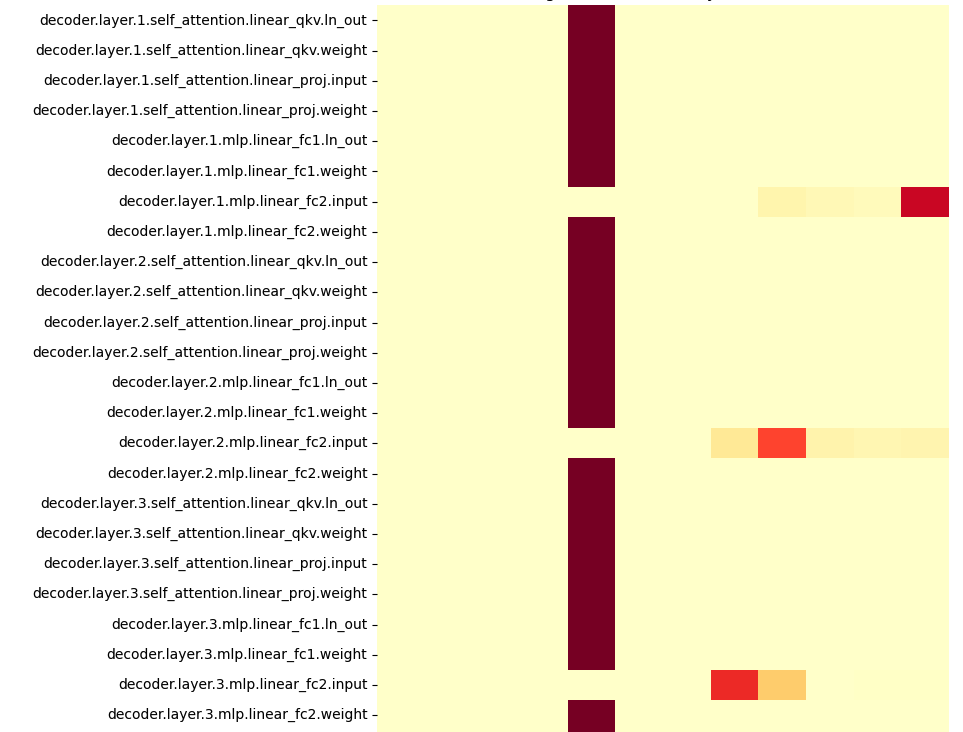}}
   
    \subfloat[Last three transformer blocks.]{\includegraphics[width=0.7\textwidth]{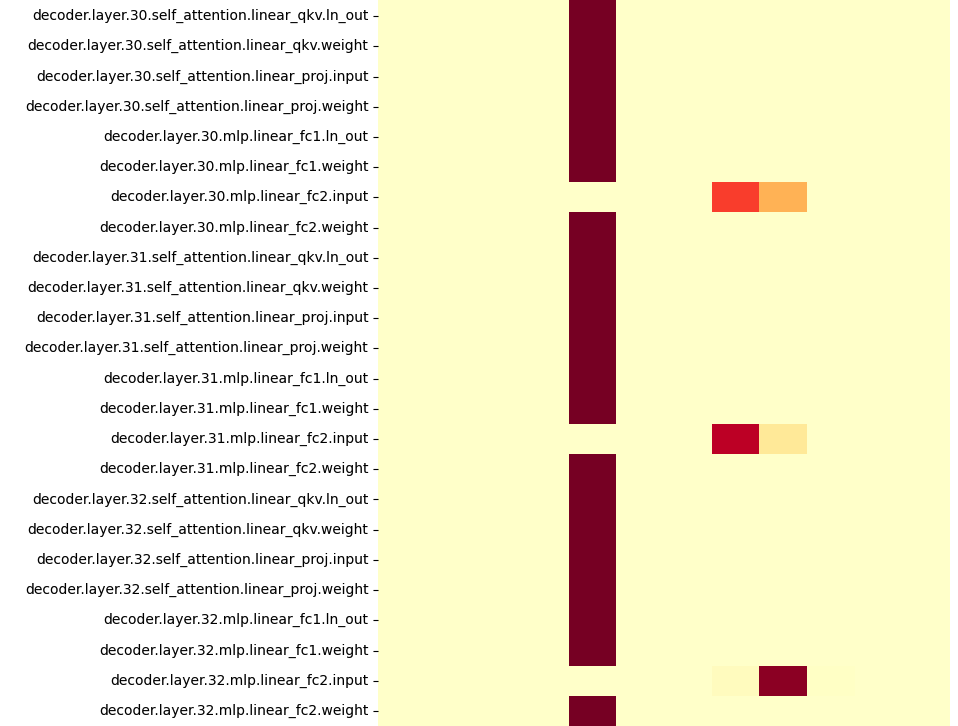}}
    
    \caption{Heatmap for the MoR block scaling algorithm in the forward pass.}
    \label{fig:block_fwd}
\end{figure}

\begin{figure}[htbp]
    \centering
    
    \subfloat[Last six transformer blocks.]{\includegraphics[width=0.7\textwidth]{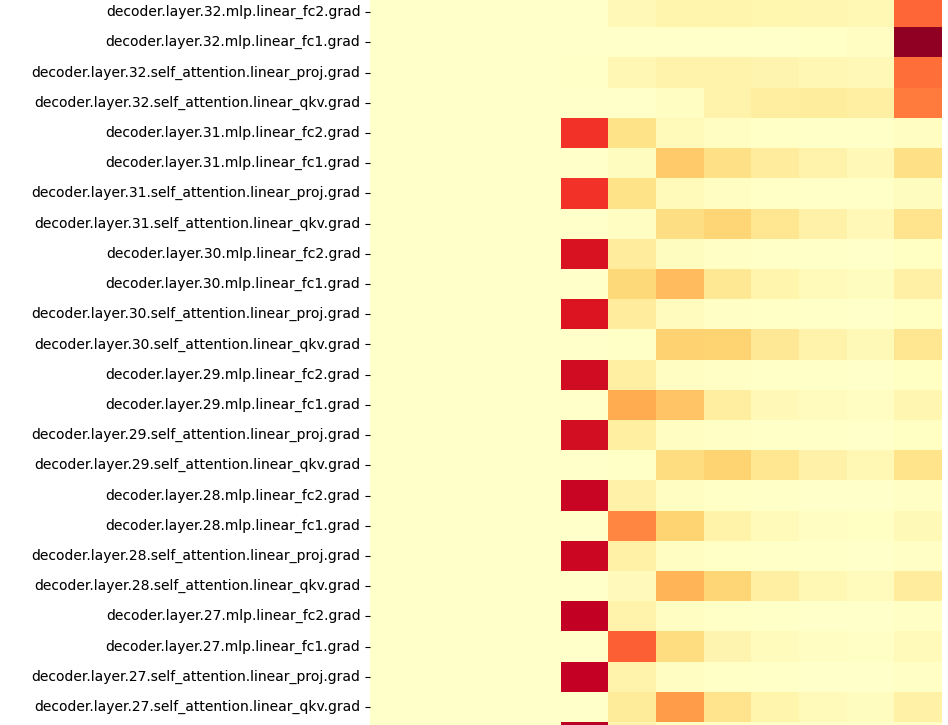}}
   
    \subfloat[First six transformer blocks.]{\includegraphics[width=0.7\textwidth]{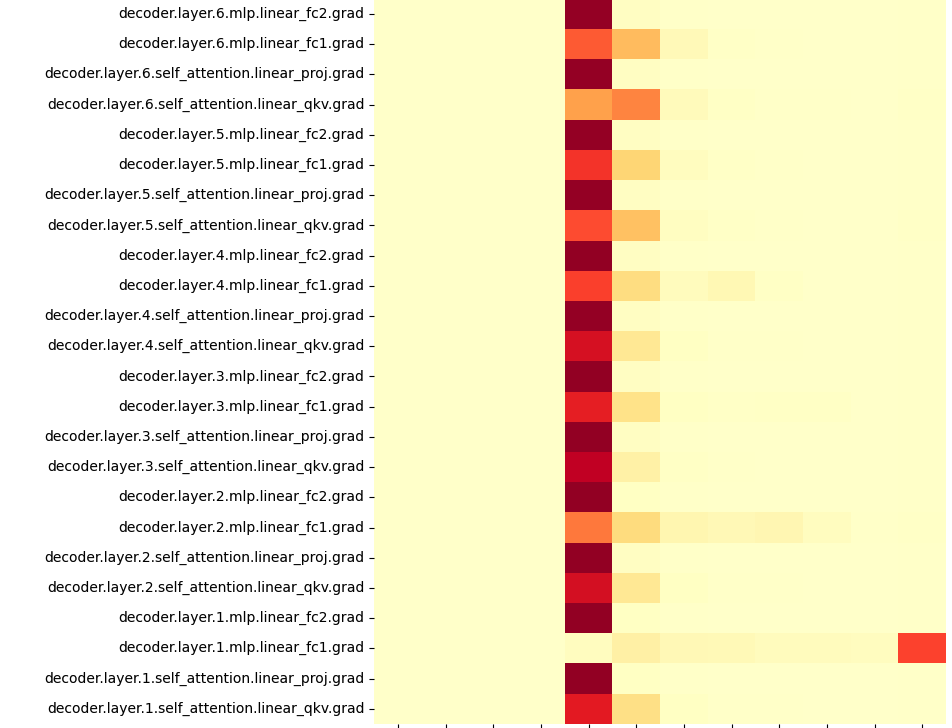}}
    
    \caption{Heatmap for the MoR block scaling algorithm in the backward pass.}
    \label{fig:block_bwd}
\end{figure}

\paragraph{Analysis of Per-Block Scaling (Configuration 1)}
We begin by analyzing the per-block strategy.
At the end of training, we can see the heatmap of the first three and last three transformer blocks 
during the forward pass in Figure \ref{fig:block_fwd}. As illustrated in the heatmap, most of the tensors
have relative error well below $2.5\%$. The only exception is the FC2 activation tensor. For the first layer, 
the majority of the counts are in the last bin, which is relative error > $5.5\%$. So MoR primarily
uses BF16 to represent this tensor. Fortunately, the histogram density shifts left when the transformer block number increases.
At the 3rd transformer block, the majority of the mass has already been below the $4.5\%$ threshold, and 
are then quantized to E4M3. For the last three layers, the FC2 activation tensor is still the tensor with higher 
relative error. But the error is within the $4.5\%$ bound. So these tensors are still quantized to E4M3.
The heatmap for the backward pass is shown in Figure \ref{fig:block_bwd}. For the last layer, all the gradient
tensors have quite high relative errors. After the last layer, the gradient tensors for the linear\_qkv module
and the fc1 module has higher relative error. But the relative error is shifting
towards left when the transformer block number decreases. So most of the gradient tensors are still quantized
to E4M3. But there is an interesting outlier at the 1st layer. The gradient tensor for the FC1 module has
quite high relative error. 

\begin{figure}[htbp]
    \centering
    \includegraphics[width=0.7\textwidth]{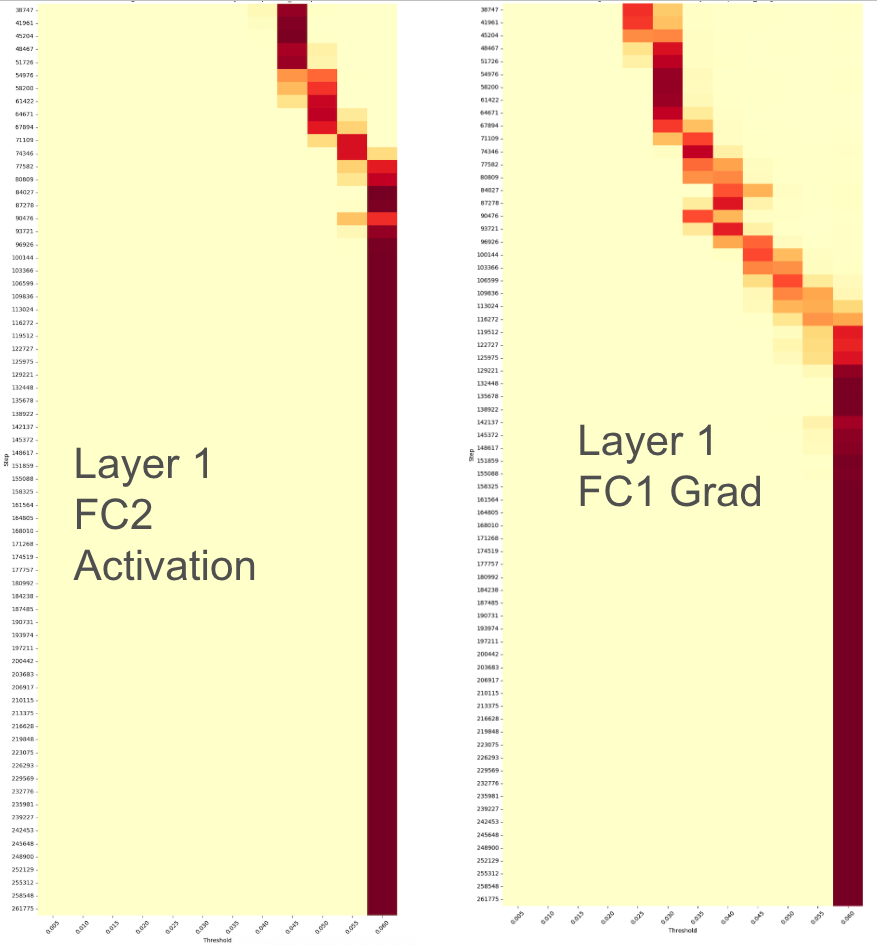}
    \caption{Heatmaps for the first transformer block. The x axis is the relative error histogram bins. The y axis is the training step.}
    \label{fig:blocklayer1}
\end{figure}

As the first layer is very special, we visualize the relative error 
based on training steps and visualize in Figure \ref{fig:blocklayer1}. 
In this heatmap, the y-axis now becomes the training step. So from the
top to bottom reveals how the relative error distribution changes when we train with more data.
A key observation is that the 
FC2 activation tensor and the FC1 gradient tensor do not start with high relative error. They start
with smaller relative error, but when training with more steps, the relative error grows and eventually
exceeds the $4.5\%$ threshold. This means that at the beginning phase of training, 
the tensors are having smaller dynamic ranges, and can be represented by E4M3 properly.
But when training make more progress, the dynamic range in the block increases over time, 
and eventually we will need to fall back to BF16 to control the quantization error. 
This ability to adapt to changing tensor statistics over time is a core strength of the MoR framework.

\begin{figure}[htbp]
    \centering
    
    \subfloat[First three transformer blocks.]{\includegraphics[width=0.7\textwidth]{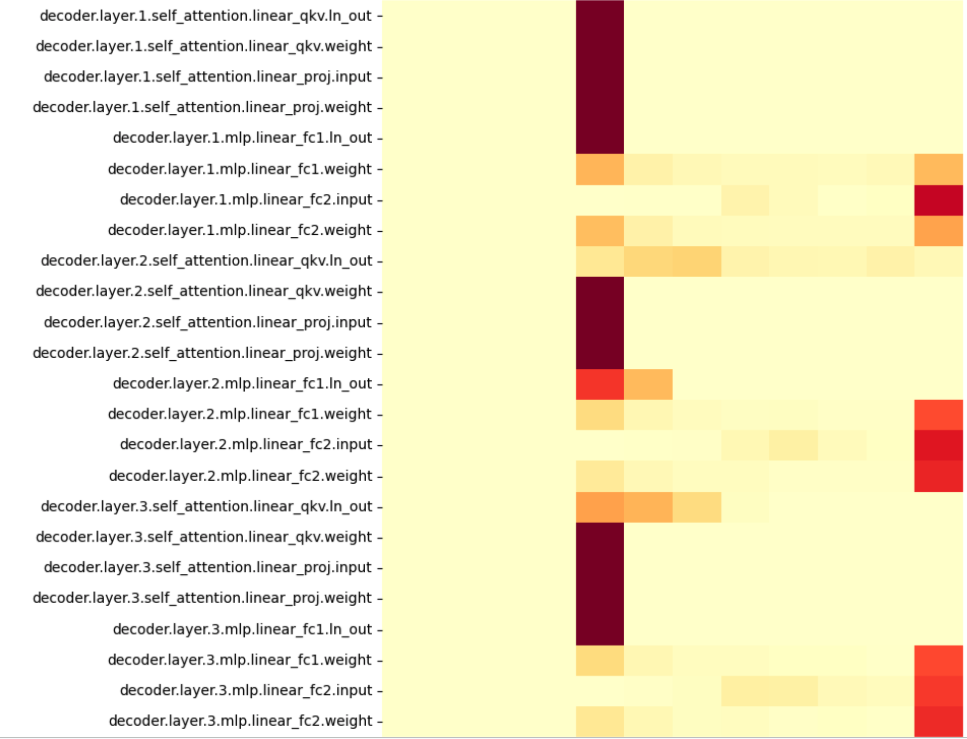}}
   
    \subfloat[Last three transformer blocks.]{\includegraphics[width=0.7\textwidth]{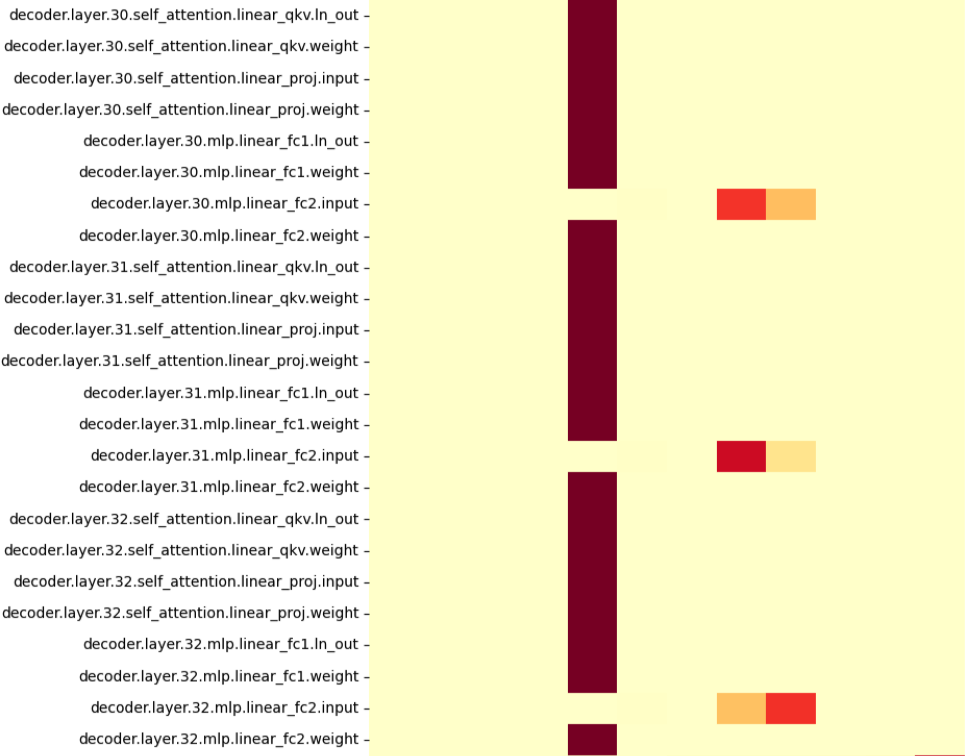}}
    
    \caption{Heatmap for the MoR block scaling algorithm in the forward pass, using the second training configuration.}
    \label{fig:train2_fwd}
\end{figure}

\begin{figure}[htbp]
    \centering
    
    \subfloat[Last six transformer blocks.]{\includegraphics[width=0.7\textwidth]{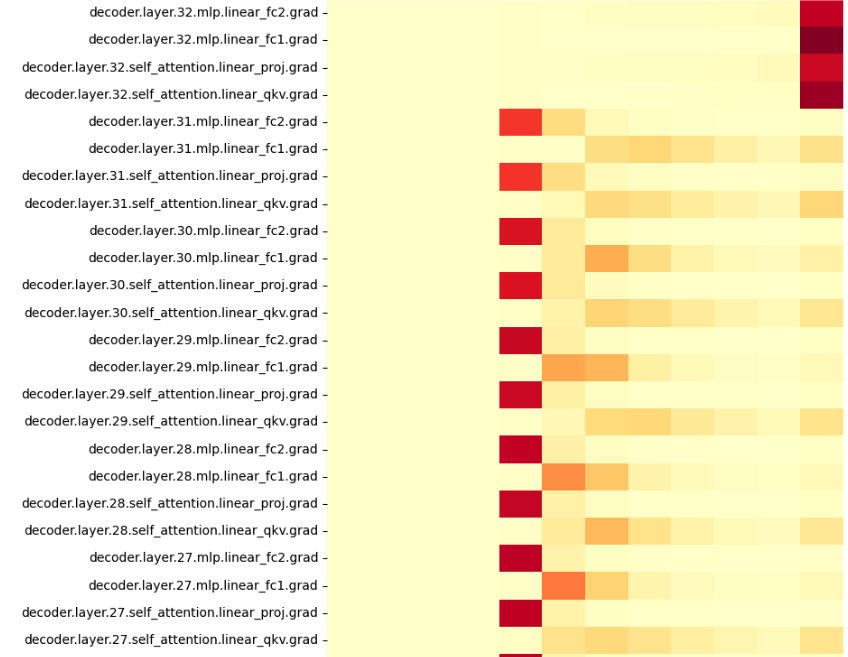}}
   
    \subfloat[First six transformer blocks.]{\includegraphics[width=0.7\textwidth]{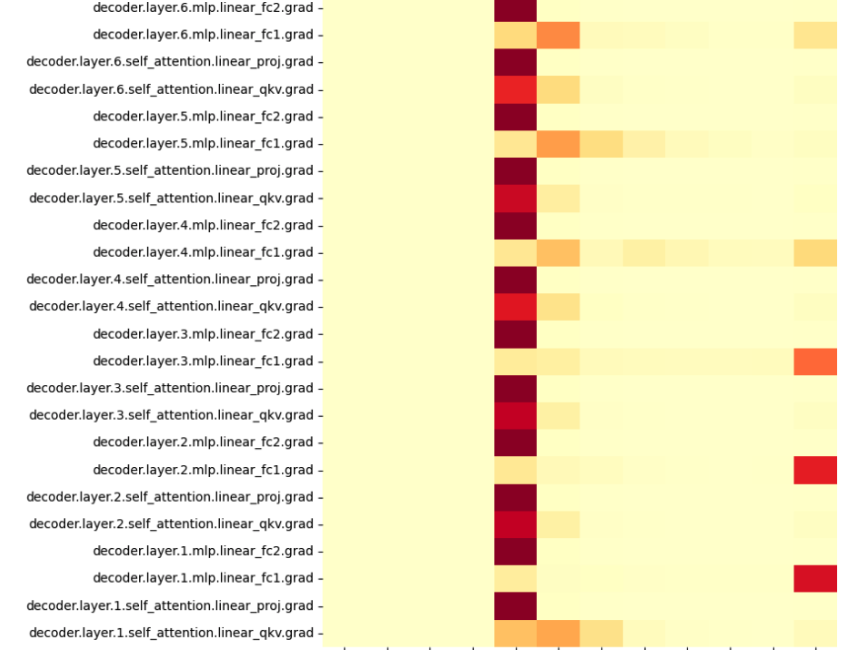}}
    
    \caption{Heatmap for the MoR block scaling algorithm in the backward pass, using the second training configuration.}
    \label{fig:train2_bwd}
\end{figure}

\paragraph{Comparing Training Configurations}
The next step we will study the differences between the first training configuration and the second training configuration
using the block scaling partition strategy. The heatmap for the first and last three transformer blocks are
summarized in Figure \ref{fig:train2_fwd}. Compared to Figure \ref{fig:block_fwd}, there are definitely more
tensors with much higher relative errors. In the heatmap, all the FC1 weight, the FC2 activation, and FC2 weight
tensors have much higher relative error in this training config. Again, there is a trend to shift towards left
with increasing the transformer block number, and there are no more BF16 fall backs after the 9th transformer block.
The heatmaps for the backward pass is shown in Figure \ref{fig:train2_bwd}. Again, the linear\_qkv and 
the FC1 gradient tensors are the ones with higher relative errors. But this time the FC1 gradient tensor has
higher relative error for all the first 6 layers instead of just the 1st layer as in the first training config.
This explains why we see the BF16 percentage has increased from $2.62\%$ to $6.38\%$ when we change the training
config from the first config to the second config.
The results demonstrate that training dynamics are highly dependent on the dataset and hyperparameters, 
and that MoR successfully adapts its quantization decisions to these different conditions.

\begin{figure}[htbp]
    \centering
    
    \subfloat[Middle three transformer blocks in the forward pass.]{\includegraphics[width=0.7\textwidth]{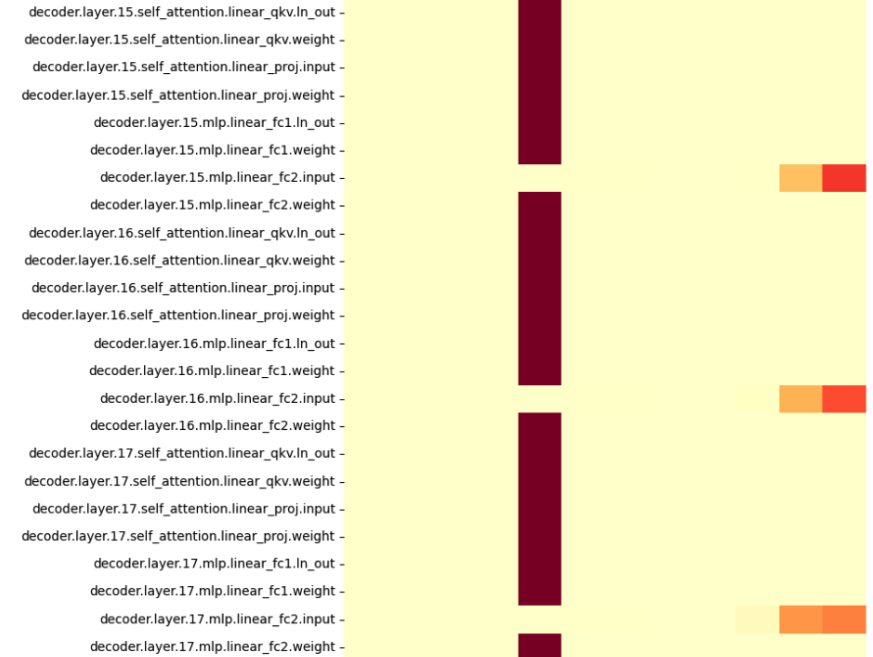}}
   
    \subfloat[Middle six transformer blocks in the backward pass.]{\includegraphics[width=0.7\textwidth]{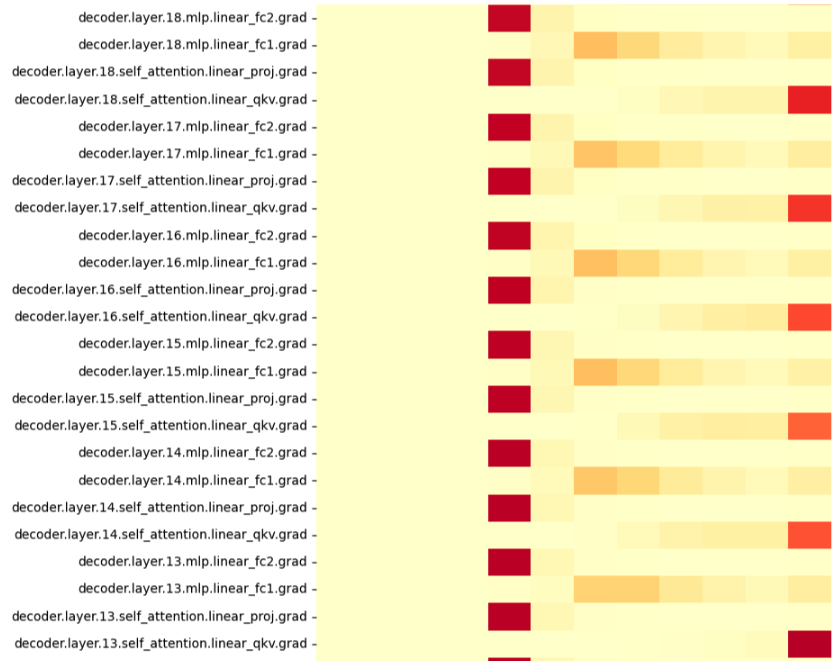}}
    
    \caption{Heatmap for the MoR tensor scaling algorithm, using the first training configuration.}
    \label{fig:current_mid}
\end{figure}

\paragraph{Comparing Partitioning Strategies}
The heatmaps also explain the quantization decision making differences between partitioning strategies. 
The \textbf{per-tensor} strategy has the highest BF16 fallback rate because its single scaling factor must cover 
the dynamic range of the entire tensor. As long as there are $2.4\%$ values that
are flushed to zero, then the relative error for the tensor will exceed $4.5\%$ and will fall back
to BF16. So it makes sense for the per-tensor partition strategy to have much higher BF16
fall back ratios. The heatmap can show exactly which tensors are now falling back to BF16.
As shown in Figure \ref{fig:current_mid}, the additional tensors that fall back to BF16 are
the tensors in the middle layers. In the forward pass, all the FC2 activations from all layers
fall back to BF16. Similarly, the gradient tensors for the linear\_qkv and fc1 modules also have
increased falling back rate in all layers. This is why the amount of BF16 tensors is much larger
for the tensor scaling partition strategy compared to the block scaling strategy.

\begin{figure}[htbp]
    \centering
    \includegraphics[width=0.7\textwidth]{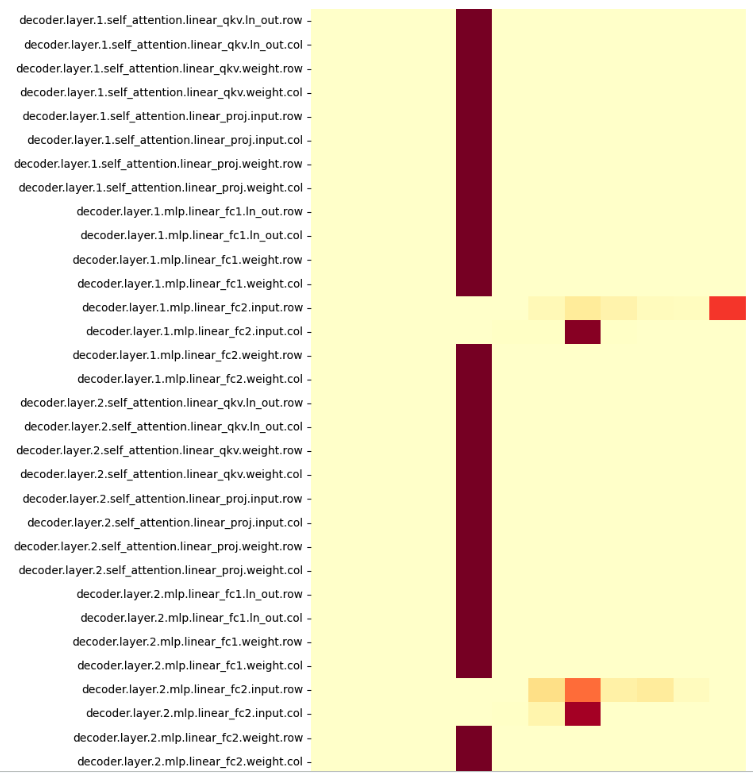}
    \caption{First two transformer blocks in the forward pass.}
    \label{fig:channel_fwd}
\end{figure}

\begin{figure}[htbp]
    \centering
    
    \subfloat[Last four transformer blocks.]{\includegraphics[width=0.6\textwidth]{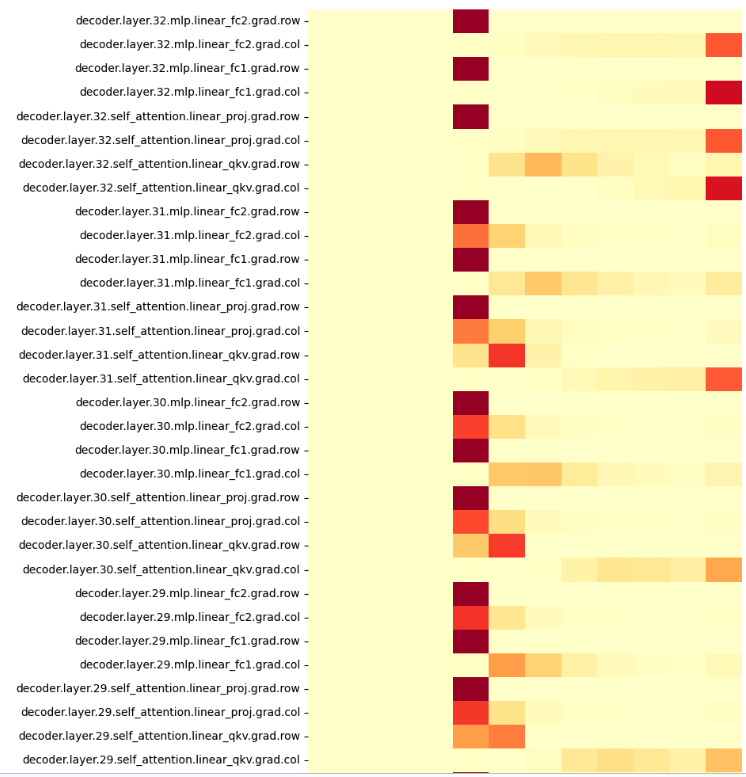}}
   
    \subfloat[First four transformer blocks.]{\includegraphics[width=0.6\textwidth]{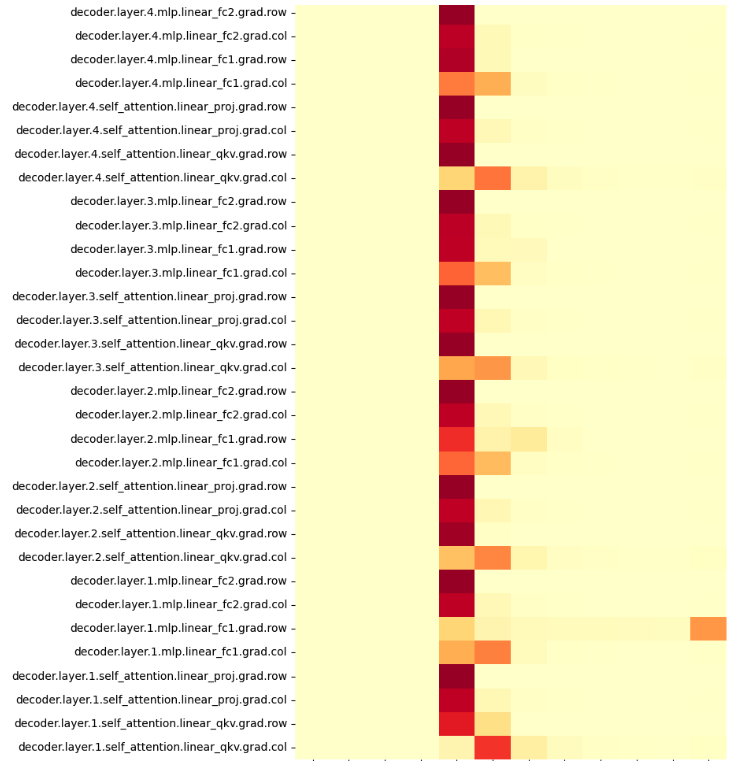}}
    
    \caption{Heatmap for the MoR channel scaling algorithm in the backward pass, using the first training configuration.}
    \label{fig:channel_bwd}
\end{figure}

Conversely, the per-channel strategy is the most efficient. 
The forward heatmap for the first two transformer blocks is in Figure \ref{fig:channel_fwd}, and 
the backward heatmaps for the first and last four transformer blocks are in Figure \ref{fig:channel_bwd}.
As illustrated in the heatmap, each tensor needs to be partitioned and quantized based on rows and columns 
according to the dot product dimension, and the relative error can differ dramatically depending on the partitioning direction. 
In the forward pass, the only tensor that needs BF16 for the per-block 
strategy is the FC2 activation tensor. When using the per-channel partition strategy, the row partition
has high relative errors, but the column partition has much smaller relative errors. So we only need BF16 for
the row partition, but not the column partition. Similarly, for all the gradient tensors in the backward pass,
one direction has much smaller relative error. 
This fine-grained selection reduces the total amount of tensors kept in BF16 by nearly half compared to the per-block strategy.

These detailed visualizations confirm that the MoR algorithm successfully adapts 
its quantization decisions based on tensor numerics at runtime, making it robust 
across different partitioning strategies and training configurations.

\subsection{Experiments for MoR at Sub-Tensor Granularity}\label{sec:exp_subtensor}

In this section, we evaluate the model quality of the sub-tensor MoR algorithms detailed in Section~\ref{sec:subtensor}.
We compare two recipes for partitioning at the $128 \times 128$ block level: the "Three-Way Selection"
algorithm, which chooses between E4M3, E5M2, and BF16, and the "Two-Way Selection" algorithm, which chooses
only between E4M3 and BF16. Both were evaluated using the first training configuration~\ref{tab:train_configs}.

\begin{figure}[htbp]
    \centering 
    
    \subfloat[Training Loss.]{\includegraphics[width=1.0\textwidth]{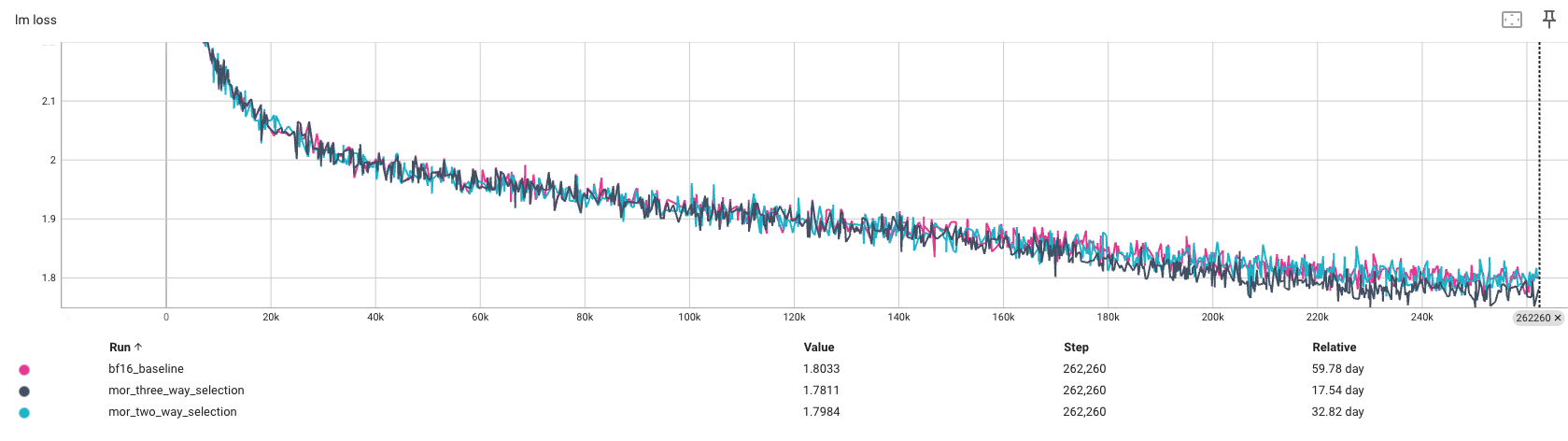}}
    
    \subfloat[Validation Loss.]{\includegraphics[width=1.0\textwidth]{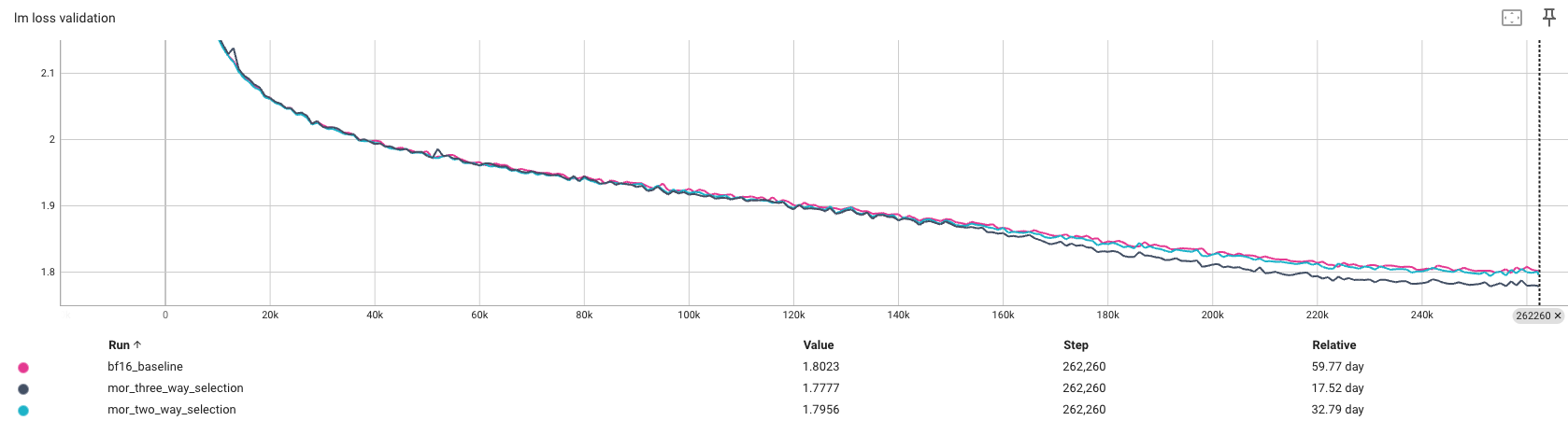}}
    
    \subfloat[L2 Norm of the Parameters.]{\includegraphics[width=1.0\textwidth]{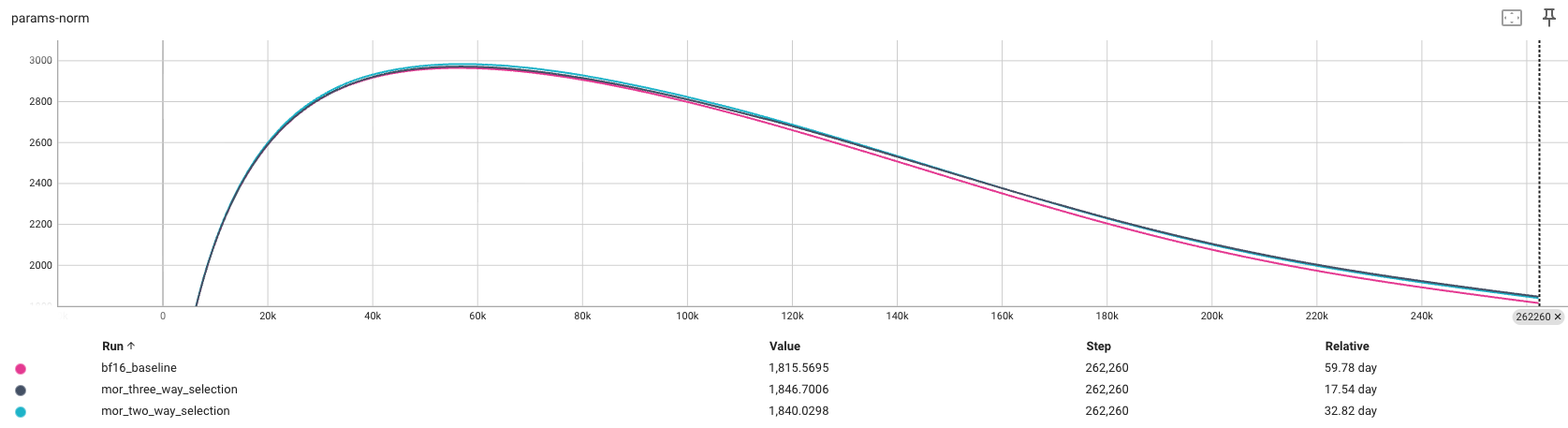}}
    
    \caption{Training loss, validation loss, and the parameter L2 Norm using the first training configuration.}
    \label{fig:train_morsubtensor}
\end{figure}

The training dynamics are presented in Figure~\ref{fig:train_morsubtensor}. While the Two-Way Selection
algorithm's loss curves closely track the BF16 baseline, the Three-Way Selection variant surprisingly
achieves a slightly better final training and validation loss.

\begin{figure}[htbp]
    \centering
    \includegraphics[width=0.7\textwidth]{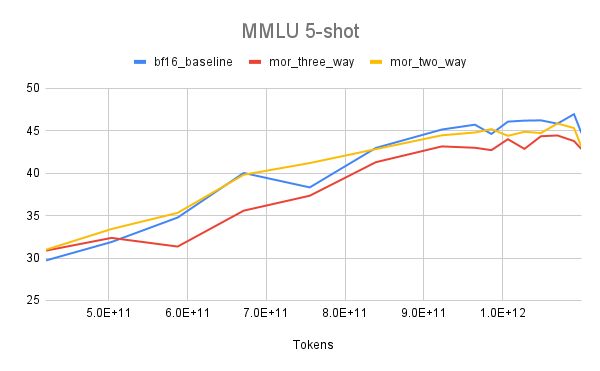}
    \caption{MMLU 5-shot scores for MoR subtensor algorithms.}
    \label{fig:mmlu_mor_subtensor}
\end{figure}

However, this apparent advantage does not translate to downstream task evaluation scores, 
as shown in Table~\ref{tab:mor_subtensor_eval}
and Figure~\ref{fig:mmlu_mor_subtensor}. The Two-Way Selection algorithm achieves results that are on-par
with, and on some tasks superior to, the BF16 baseline. In contrast, the Three-Way Selection algorithm exhibits
a significant quality degradation on most downstream benchmarks, which is particularly evident in the MMLU
performance curve.

The divergence between the validation loss and downstream task evaluation scores suggests that the Three-Way Selection algorithm
overfits to the training and validation data distributions. While the additional flexibility of the E5M2 format
helps minimize loss on in-distribution data, it appears to harm the model's ability to generalize to the different
data distributions of the evaluation benchmarks. Investigating the precise mechanism by which the inclusion of
E5M2 leads to this overfitting behavior will be continued in future works.

\begin{table}[htbp]
    \centering
    \caption{Model quality comparison for different MoR subtensor algorithms.}
    \label{tab:mor_subtensor_eval}
    \begin{tabular}{l ccc}
        \toprule
        \textbf{Metric} & \textbf{BF16} & \textbf{Two-Way Selection} & \textbf{Three-Way Selection} \\
        \midrule
        Training Loss    & 1.8033 & 1.7984 & \textbf{1.7811} \\
        Validation Loss  & 1.8023 & 1.7956 & \textbf{1.7777} \\
        \midrule
        MMLU             & 44.72  & \textbf{46.56}  & 42.84  \\
        WinoGrande       & 66.69  & \textbf{68.35}  & 66.61  \\
        PIQA             & 78.45  & \textbf{78.51}  & 78.24  \\
        HellaSwag        & \textbf{74.93}  & 74.67  & 73.58  \\
        Arc-Easy         & 73.48  & \textbf{74.24}  & 71.13  \\
        Arc-Challenge    & 41.30  & 41.30  & \textbf{41.89}  \\
        OpenBookQA       & \textbf{42.80}  & 40.20  & 41.60  \\
        SIQA             & 44.63  & \textbf{45.70}  & 44.01  \\
        CommonSenseQA    & 34.32  & \textbf{40.87}  & 29.07  \\
        \bottomrule
    \end{tabular}
\end{table}

\clearpage
\section{Conclusions}\label{sec:conclusions}

In this paper, we introduced two primary contributions to the field of mixed-precision training. 
First, we proposed the Group Amax Mantissa (GAM) scaling algorithm, a novel method that accurately 
preserves the absolute maximum value of a tensor group, enhancing the precision of the scaling process. 
Second, we developed the Mixture-of-Representations (MoR) framework, a dynamic, metric-driven approach to 
selecting quantization formats at runtime.

Our experiments demonstrate that the tensor-level MoR algorithm is remarkably robust and effective. 
Using a single, empirically-derived relative error threshold of 4.5\%, our method achieved on-par 
model quality with a standard BF16 baseline across three different partitioning strategies and two distinct training configurations. 
The approach is also highly efficient; our best-performing per-channel strategy successfully quantized over $98.38\%$/$95.93\%$ of 
tensors to E4M3 in the two training configurations, requiring only a small fraction to remain in high precision.

Furthermore, we presented preliminary but promising results for sub-tensor MoR. 
The two-way (E4M3/BF16) selection recipe also maintained model quality comparable to the baseline. 
While our investigation of this finer granularity is ongoing, we believe the flexibility of the sub-tensor approach 
holds significant potential for future work aimed at further optimizing the memory and compute footprints of large model training.

\section*{Acknowledgments}
The authors wish to thank Dusan Stosic, Oleg Rybakov, Mohammad Shoeybi, Bryan Catanzaro, Rama Govindaraju, Mike Houston, Misha Smelyanskiy, Carlo del Mundo, Eric Chung, and Jonah Alben for their support and valuable feedback throughout this research.

\bibliographystyle{unsrtnat}
\bibliography{references}  






\end{document}